\documentclass[10pt,twocolumn,letterpaper]{article}

\usepackage{iccv}
\usepackage{times}
\usepackage{epsfig}
\usepackage{graphicx}
\usepackage{amsmath}
\usepackage{amssymb}

\usepackage{color}
\usepackage{xcolor}
\usepackage{epsfig}
\usepackage{subfig}
\usepackage{multirow}
\usepackage{booktabs}    
\usepackage{comment}

\usepackage[pagebackref=true,breaklinks=true,letterpaper=true,colorlinks,bookmarks=false]{hyperref}

\newcommand{\real}{\mathbb{R}}

\newcommand{\figref}[1]{Fig.~\ref{#1}}
\newcommand{\tabref}[1]{Table~\ref{#1}}
\newcommand{\secref}[1]{Sec.~\ref{#1}}
\newcommand{\eqnref}[1]{Eqn.~\ref{#1}}

\iccvfinalcopy 

\def\iccvPaperID{6271} 

\ificcvfinal\fi

\begin{document}

\title{
Unsupervised Depth Completion with Calibrated Backprojection Layers
}

\author{Alex Wong \\
UCLA Vision Lab \\
{\tt\small alexw@cs.ucla.edu}
\and
Stefano Soatto \\
UCLA Vision Lab \\
{\tt\small soatto@cs.ucla.edu }
}

\maketitle
\ificcvfinal\thispagestyle{empty}\fi

\begin{abstract}
We propose a deep neural network architecture to infer dense depth from an image and a sparse point cloud. It is trained using a video stream and corresponding synchronized sparse point cloud, as obtained from a LIDAR or other range sensor, along with the intrinsic calibration parameters of the camera. At inference time, the calibration of the camera, which can be different than the one used for training, is fed as an input to the network along with the sparse point cloud and a single image.   
A Calibrated Backprojection Layer backprojects each pixel in the image to three-dimensional space using the calibration matrix and a depth feature descriptor. The resulting 3D positional encoding is concatenated with the image descriptor and the previous layer output to yield the input to the next layer of the encoder. A decoder, exploiting skip-connections, produces a dense depth map. The resulting Calibrated Backprojection Network, or KBNet, is trained without supervision by minimizing the photometric reprojection error. KBNet imputes missing depth value based on the training set, rather than on generic regularization. We test KBNet on public depth completion benchmarks, where it outperforms the state of the art by 30.5\% indoor and 8.8\% outdoor when the same camera is used for training and testing. When the test camera is different, the improvement reaches 62\%. Code available at: \url{ https://github.com/alexklwong/calibrated-backprojection-network}.
\end{abstract}

\section{Introduction}
Sensor platforms designed to enable interaction with physical space often include optical as well as range sensors. From cars to phones, cameras are paired with active sensors such as LIDARs, Sonars or Radars. We address the case of a single camera and a single sensor that returns the three-dimensional (3D) coordinates of a number of points far fewer than the number of pixels in the RGB image. The range sensor alone provides a sparse estimate of the Euclidean geometry of the surrounding environment, but often insufficient for planning in applications such as autonomous navigation or manipulation. We wish to leverage the complementarity of the optical and range modalities to provide a \textit{dense} depth map, whereby a range value\footnote{The depth associated with the pixel is the Euclidean distance of the closest point in the scene along the projection ray through that pixel and the optical center. We assume the sensors to be calibrated and synchronized, and in particular the intrinsic calibration matrix of the camera is known so that pixel coordinates can be converted to Euclidean 3D coordinates.} is associated to every pixel in the image (in the millions) as opposed to just the LIDAR or radar returns (in the thousands). 

\textit{Depth completion} consists of mapping a single RGB image and a sparse 3D point cloud onto a dense depth map, which requires inferring a depth value where missing. This can be done by means of regularization, or inductively using previously observed data for scenes other than the present. We assume we have available a \textit{training set} consisting of monocular videos, corresponding sparse 3D point cloud, and intrinsic calibration matrix of the camera used for capture,\footnote{Typically, range and optical sensors are calibrated mechanically and pre-registered, so extrinsic calibration is not needed.} but without any manual annotation or ground-truth dense depth i.e. \textit{unsupervised}.

Our goal is to use the training set to learn a function that, for a scene and camera not used for training, can map a sparse point cloud registered to an image, along with the matrix of intrinsic calibration parameters of the camera, and produce a dense depth map associated with the test image. 

We propose a novel deep neural network architecture that leverages a \textit{sparse-to-dense} (S2D) module and \textit{calibrated backprojection} (KB) layers. S2D is comprised of various pooling and convolutional layers to yield a dense representation of the sparse points. A KB layer then maps camera intrinsics, input image, and current depth estimate onto the 3D scene. This can be thought of as a form of \textit{spatial (Euclidean) positional encoding} of the image. Unlike previous architectures, camera intrinsics are an \textit{input} to our model, as opposed to a fixed set of parameters in the training loss. This allows us more flexibility to transfer the trained model to sensor platforms other than that used for training. 

Our network is trained unsupervised with the standard Photometric Euclidean Reprojection Loss (PERL) i.e. the absolute difference between a reconstructed image and the actual image measured at a time instant. We also penalize the reconstruction error of the input sparse points and Total Variation of the estimated depth map, a standard sparsity-inducing prior to reduce the penalty for large depth changes at adjacent pixels that straddle occluding boundaries. At test time, no video is necessary and inference is performed on each image and sparse point cloud independently.

These innovations allow us to improve the baseline \cite{wong2020unsupervised} and state of the art \cite{wong2021learning} by an average of 13.7\% and 8.8\%, respectively, on outdoors (KITTI \cite{uhrig2017sparsity}), and 51.7\% and 30.5\% indoors (VOID \cite{wong2020unsupervised}), when calibration is the same for training and testing. When different calibrations are used, our method generalizes better than the baseline and state of the art by 83\% and 62\%, respectively, in relative error. All of this is achieved with a smaller computational footprint thanks to the inductive bias induced by KB layers, which allows us to use a smaller network than current methods.

\subsection{Related Work and Contributions}
\label{sec:related_work}
Depth completion is a form of imputation, which requires regularization that hinges the assumption that \textit{``nearby points''} should be
assigned ``similar'' (depth) values. Methods differ in the choice of topology i.e. what points should be considered ``nearby,'' and how to combine the values of such points to impute the missing depth value. 

\textbf{Generic Image-Based Regularization.} In image topology, nearby points correspond to adjacent pixels. This is not a good choice, for their depths can be arbitrarily different at occluding boundaries. In image segmentation, the RGB values are used to define a topology to partition the image domain into connected regions of nearby points, putatively corresponding to ``objects.'' The topology induced by (color) values can be exploited by minimizing Total Variation (TV \cite{rudin1992nonlinear} and ``Color TV'' \cite{blomgren1998color}) while trying to reproduce the image itself. We adopt TV as a generic regularizer since the statistics of natural range images are very similar to that of natural (intensity) images \cite{mumford1989optimal}, whereby the gradient distribution is highly kurtotic, corresponding to homogeneous smooth regions separated by sharp boundaries. 

\textbf{Data-driven Regularization.} 
``Closeness'' among pixels can be defined not just within the same image, but across different images in the training set. In this case, the regularity criterion is not explicit, but implicit in the inductive bias used for training. Before training starts, the bias is encoded in the training loss ($L^1$ prediction error), the generic regularizers (TV), the training set, and the choice of architecture and optimization. After training is completed, all these biases are burnished in the parameters (weights) of the trained model, which inform the prediction of our depth map and therefore act as a regularizing mechanism. 

Among data-driven methods for depth completion, many are \textbf{supervised}. Early works cast depth completion as compressive sensing \cite{chodosh2018deep} and as morphological operators \cite{dimitrievski2018learning}.  Recent works focused on network operations \cite{eldesokey2018propagating,huang2019hms} and architectures \cite{chen2019learning,ma2019self,uhrig2017sparsity,yang2019dense} to effectively deal with the sparse inputs. \cite{ma2019self} proposed an early fusion architecture while \cite{jaritz2018sparse,yang2019dense} used late fusion to process each data modality separately. \cite{huang2019hms} performed joint concatenation and convolution to upsample the sparse depth. \cite{chen2019learning} proposed a 2D-3D fusion network while \cite{li2020multi} used a cascade hourglass network. \cite{cheng2020cspn++} used a convolutional spatial propagation network and \cite{park2020non} leveraged non-local spatial propagation. Whereas, \cite{eldesokey2018propagating,eldesokey2020uncertainty,qu2021bayesian,qu2020depth} learned uncertainty of estimates, \cite{van2019sparse} leveraged  confidence maps, and \cite{ qiu2019deeplidar,xu2019depth,zhang2018deep} used surface normals for guidance. Like us, \cite{merrill2021robust,sartipi2020deep,zuo2021codevio} proposed light-weight networks suitable for use with SLAM/VIO systems.

All of these methods require ground truth for training, which is often unavailable and, when available, prohibitively expensive \cite{uhrig2017sparsity}. Hence, these methods are limited to offline training. But even if ground truth were available online, most of these methods employ complex architectures with many layers and parameters, e.g. 25.84M for \cite{park2020non}, 53.4M \cite{qiu2019deeplidar}, and 28.99M \cite{xu2019depth}, and thus are not suitable for learning online. Instead, we propose to learn dense depth from the virtually limitless amount of un-annotated images and sparse point clouds via a predictive cross-modal validation criterion. Our proposed architecture only uses 6.9M parameters and our choice of supervision allows us to continuously learn even after the system is deployed. 

\textbf{Unsupervised/Self-supervised depth completion} assumes stereo images or monocular videos to be available during training. Both stereo \cite{shivakumar2019dfusenet,yang2019dense} and monocular \cite{ma2019self,wong2021learning,wong2021adaptive,wong2020unsupervised} training paradigms leverage sparse depth reconstruction and photometric reprojection error as a training signal by minimizing photometric discrepancies between the input image and its reconstruction from other views. \cite{ma2019self} used Perspective-n-Point \cite{lepetit2009epnp} and RANSAC \cite{fischler1981random} to align consecutive video frames. However, \cite{ma2019self} does not generalize well to indoor scenes with many textureless surfaces. \cite{yang2019dense} learned a depth prior conditioned on the image by pretraining a separate network on ground truth from an additional dataset. As mentioned earlier, this is not scalable; also, using a network trained on a specific domain (e.g. outdoors) as supervision will not generalize (e.g. indoors). Unlike \cite{yang2019dense}, our method does not require ground truth and is not limited to a specific domain. \cite{lopez2020project,wong2021learning} leverage additional synthetic datasets, which require dealing with sim-to-real; our method is able to achieve the state-of-the-art \textit{without} needing access to additional data.

The challenge of depth completion is precisely the sparsity, which renders convolutions ineffective as the activations of early layers tend to be zeros as well. To obtain a denser representation, early layers must propagate (or densify) the signal. As a result, \cite{ma2019self,shivakumar2019dfusenet,yang2019dense} employed very deep networks with many layers and parameters in order to learn the map from sparse depth and image to dense depth. To handle this problem, \cite{wong2020unsupervised} approximated the scene with a hand-crafted mesh, but it is not differentiable and prone to errors in regions with very few points or complex structures. \cite{wong2021learning} proposed spatial pyramid pooling (SPP), but their max pooling layers decimated details on closer objects. Instead, we propose a fully differentiable sparse-to-dense module that learns the trade-off between density and detail to retain both near and far structures. 

Our work goes counter to the trend of forgoing inductive bias, i.e. learning everything with generic architectures like Transformers \cite{vaswani2017attention}, including what we already know such as basic Euclidean geometry. Our model has a strong inductive bias in our calibrated backprojection layer, which incorporates the calibration matrix directly into the architecture to yield an RGB representation lifted into scene topology via 3D positional encoding. This may seem futile as we could just add intrinsics to the long list of parameters to be learned \cite{gordon2019depth}. However, unlike semantic retrieval, spatial inference requires \textit{identifiability}: There is \textit{one} true scene in front of us, and unless information about calibration is available and properly exploited, inference yields one of infinitely many depth maps that are equally good at predicting the next frame in the training set. Since there is no supervision, calibration mediates the relation between the prediction error and the \textit{true} depth. Because existing methods use calibration in the computation of the loss, which the intrinsics are encoded in the weights, hampering transferability. In our architecture, calibration is an input, which can be changed at inference time. While one could pre-process the images to a canonical calibration, this introduces latency, cost and artifacts that can affect the reconstruction quality. We note that \cite{handa2016gvnn,riba2020kornia} proposed backprojection as a layer and \cite{facil2019cam} used calibration as input, but we are the first to consider an RGB 3D representation for depth completion.

\begin{figure}
    \centering
    \includegraphics[width=1.0\linewidth]{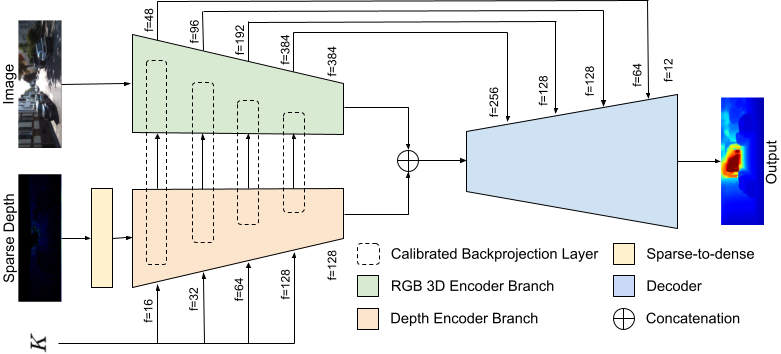}
    \caption{\textit{KBNet architecture.} Our architecture takes, as input, an RGB image, the corresponding sparse depth map and camera calibration matrix. We first learn a dense representation of the sparse points with our sparse-to-dense module. The result of which and the calibration matrix are used for calibrated lifting, which allows us to backproject image features onto 3D space (akin to spatial positional encodings) to yield a RGB 3D representation. Our network is very light-weight and fast, yet achieves the state of the art.}
    \label{fig:network_architecture}
    \vspace{-1.3em}
\end{figure}

\noindent{\bf Our contributions} include (a) a sparse-to-dense module that learns a dense representation of the sparse point cloud, (b) an unsupervised depth completion method that takes calibration information as input to the model, and (c) incorporates it directly in the architecture through a novel \textit{calibrated backprojection} module, which represents spatial positional encoding that is transferred laterally across different branches of the encoder. The resulting inductive bias helps select, among all depth, maps compatible with the prediction loss, those that result in a Euclidean (calibrated) reconstruction. The strong inductive bias allows us to (d) reduce the computational footprint, increase generalization and achieve performance beyond the state of the art despite having fewer parameters.

\section{Method Formulation}
\label{sec:method_formulation}
Our goal is to recover a 3D scene from an RGB image $I : \Omega \subset \real^2 \mapsto \real^3_+$ and the associated sparse point cloud projected onto the image plane $z : \Omega_{z} \subset \Omega \mapsto \real_+$, without access to ground-truth depth annotations. 

\begin{figure}
    \centering
    \includegraphics[width=1.0\linewidth]{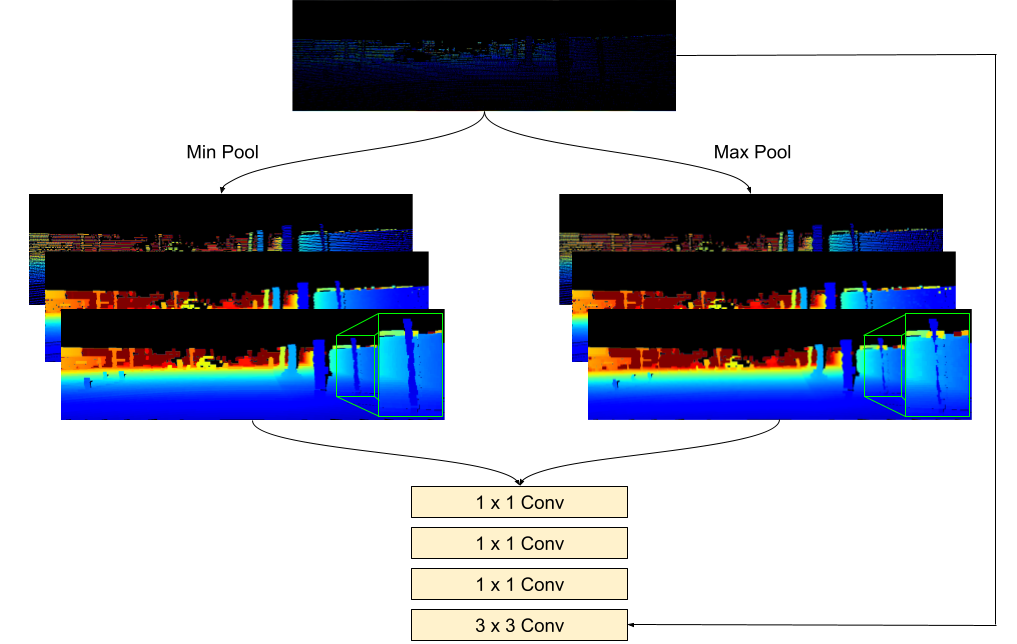}
    \caption{\textit{Sparse-to-dense module.} We perform min and max pooling with various kernel sizes to produce a dense representation. There exists trade-offs between density and detail (large vs. small kernel sizes) and preservation of near and far structures (min vs. max pooling, as highlighted in green). We balance these trade-offs with $1 \times 1$ convolutions and fuse the result with the input via a $3 \times 3$ convolution.}
    \label{fig:sparse_to_dense_module}
    \vspace{-1.3em}
\end{figure}

We propose a sparse-to-dense module (\figref{fig:sparse_to_dense_module}) $f_\omega$, parameterized by $\omega$, that captures local and global structure of the sparse inputs by combining min and max pooling at different scales. The result is a dense or quasi-dense depth representation $f_\omega(z)$, depending on the sparsity of the input, which frees the rest of network to utilize its early convolutional layers to learn scene structure rather than to densify the input -- making the overall architecture more efficient.

The sparse-to-dense module (\secref{sec:sparse_to_dense_module}) is part of an overall encoder-decoder architecture $f_\theta$, parameterized by $\theta$, called KBNet (\secref{sec:network_architecture}), that includes a Calibrated Backprojection layer which explicitly backprojects pixels onto 3D space using intrinsic camera calibration and depth encoding from $f_\omega$ . Unlike previous works \cite{ma2019self,shivakumar2019dfusenet,wong2021learning,wong2020unsupervised,yang2019dense} that encode depth and image in two separate branches, we leverage camera calibration and our depth encoding to lift the image representation to 3D and passed it to the decoder via skip connections. KBNet  (\figref{fig:network_architecture}) 
produces dense depth $\hat{d} := f_\theta(f_\omega(z), I, K)$, where $K \in {\mathbb R}^{3\times 3}$ is the upper-triangular matrix of intrinsic calibration parameters. To train our model, we use monocular videos to compose a loss function from temporally adjacent frames (\secref{sec:loss_function}).

\subsection{Sparse-to-Dense Module (S2D)}
\label{sec:sparse_to_dense_module}
Our S2D module $f_{\omega}$ (\figref{fig:sparse_to_dense_module}) performs multi-scale densification on the input sparse depth map $z$ using a series of min and max pooling layers with various kernel sizes, which are chosen based on the sparsity of the point cloud e.g. from LIDAR returns or sparse points tracked by VIO \cite{fei2019geo} (see Supp. Mat. for kernel sizes). The outputs of the pooling layers are concatenated and fed into three $1 \times 1$ convolutions to learn the trade-offs between pooling types and kernel sizes. The result of which is fused with the $z$ via a $3 \times 3$ convolutional layer, yielding a dense or quasi-dense depth representation that is fed to the rest of the network $f_\theta$.

Because the depth inputs are sparse, we design our min pooling layers to avoid pooling zeros or invalid (negative) depth values. We set all values $z(x)$ less than zero to be infinity for $x \in \Omega$:
\begin{equation}
    z'(x) = \begin{cases} 
        z(x) &\mbox{if } z(x) > 0 \\
        \infty & \mbox{otherwise}.   
    \end{cases}
\end{equation}
$z'$ is fed to a min pooling layer with $k \times k$ kernel size,
\begin{equation}
    p = \texttt{minpool}(z', k).
\end{equation}
Finally, for all $x$, any infinity values pooled due to large empty regions are set to zero:
\begin{equation}
    p_{min}(x) = \begin{cases} 
        p(x) &\mbox{if } p(x) \neq \infty \\
        0 & \mbox{otherwise}.
    \end{cases}
\end{equation}
Our approach involves two main trade-offs: (i) density versus detail and (ii) preservation of near versus far structures.

\textbf{Density versus details.} For the purpose of densification, one may perform pooling with large kernel sizes, but it comes at the expense of details of local structures. In contrast, pooling with small kernel sizes in an attempt to retain detail will result in very few neuron activations, which hinders learning. Hence, to retain local details while obtaining a dense representation, we propose to perform pooling with both small and large kernel sizes.

\textbf{Near versus far.} When pooled solely with max pooling, farther structures are preserved, but details of the closer ones are decimated as the kernel size grows larger. For instance in \figref{fig:sparse_to_dense_module}, thin structures close to the camera i.e. the highlighted pole ``disappears'' due to large max pooling kernel size. On the other hand, when only using min pooling, the closer structures become more prominent, but in turn, the farther regions are corrupted. Moreover, in cluttered scenes, min pooling causes adjacent structures to ``bleed'' into each other. Hence, to preserve close and far structures, we employ both min and max pooling layers.  

To optimize for both trade-offs, we concatenate the outputs of min and max pooling together and feed them into $1 \times 1$ convolutional layers. Finally, we use a $3 \times 3$ convolution to fuse the multi-scale depth features back into the original sparse depth via a residual connection, yielding a dense representation $f_\omega(z)$ to be fed to $f_\theta$. 

We note that our S2D bares some resemblance to spatial pyramid pooling (SPP) \cite{he2015spatial}; however, SPP was designed to ensure the same size feature maps are maintained when different size of inputs. It is also intended to operate on dense inputs. While \cite{wong2021learning} also proposed an SPP for sparse inputs, its use of max pooling decimated details for nearby structures. Neither are substitutes for our S2D module. 
 
\subsection{KBNet Architecture}
\label{sec:network_architecture}
\textbf{Motivation.} Unsupervised methods \cite{ma2019self,wong2021learning,wong2021adaptive,wong2020unsupervised} use the photometric reprojection error $\ell_{perl}$ as a training signal. The input image $I_t$ is reconstructed from temporally adjacent frames $I_{\tau}$ for $\tau \in T \doteq \{t-1, t+1\}$ to yield $\hat{I}_\tau$, 
\begin{equation}
\label{eqn:image_reconstruction}
    \hat{I}_\tau(x, \hat{d}, g_{\tau t}) = I_{\tau} \big( \pi  g_{\tau t} K^{-1} \bar{x} \hat d(x) \big),
\end{equation}
and the per pixel photometric reprojection error is measured by $\ell_{perl} = |\hat{I}_{\tau}(x, \hat d, g_{\tau t})-I_{t}(x)|$.
Here $\bar{x} = [x^\top, 1]^\top$ are the homogeneous coordinates of $x\in\Omega$. Using the notation in \cite{ma2012invitation},  $g_{\tau t} \in SE(3)$ is the relative pose (rotation and translation) of the camera from time $t$ to time $\tau$, $K$ denotes the intrinsic calibration matrix, and $\pi$ is a canonical perspective projection. For simplicity, we  will refer to the reconstruction from time $\tau$ at a coordinate $x$ as $\hat{I}_\tau(x)$.

Inferring Euclidean structure and motion in the absence of calibration information is notoriously difficult and dependent on conditions rarely satisfied in ordinary training videos, such as rotation around three independent axes \cite{ma2012invitation}. Minimizing any form of $\ell_{perl}$ forces the network to implicitly learn the calibration matrix $K$, as all prior work does. As pretrained models are commonly deployed on sensor platforms different than those used during training, this hinders generalization as the network becomes overfitted to the camera used to collect training data. In contrast, our network, KBNet, takes it as input; this allows us to use different calibrations in training and test, which significantly improves generalization (\tabref{tab:nyuv2_test_set_results}).

\begin{figure}
    \centering
    \includegraphics[width=1.0\linewidth]{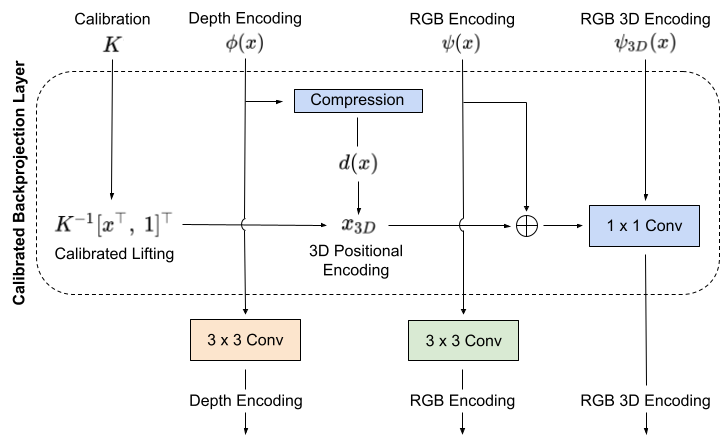}
    \caption{\textit{Calibrated Backprojection (KB) Layer.} The standard depth and color image encoding layers \cite{wong2020unsupervised} are combined using the calibration matrix as additional input. Calibration is used to lift pixel coordinates to three dimensions, which are backprojected by a compressed depth descriptor into a 3D positional encoding. The result is concatenated with the image encoding and the output of the previous KB layer, and fused with a $1 \times 1$ convolution. This yields an RGB 3D representation, which is used as a skip connection to the decoder and input to subsequent layers. 
}
\vspace{-1.3em}
\label{fig:calibrated_backprojection_layer}
\end{figure}

\textbf{Calibrated Backprojection Layers} 
take, as input, the depth and RGB image encodings, and the camera calibration matrix $K$ and output not only the corresponding encodings of the depth map and of the RGB image, but also an encoding of the RGB image backprojected onto 3D space. Once we have formed this RGB 3D representation, it is fed as input to subsequent Calibrated Backprojection (KB) layers and as skip connection to the decoder and once we have form this representation (\figref{fig:calibrated_backprojection_layer}).

To realize a KB layer, first, we use the calibration matrix to lift the coordinates of each pixel $x \in \Omega$ to three dimensional space $x \rightarrow K^{-1} \bar{x}$. Then, the feature map of the depth encoder $\phi(x) \in \real^M$, with $M$ ranging from 16 in the first layer to 128 in the last one, is collapsed to a scalar by a trainable projection or ``compression'' module $q$, $d(x) = q^\top \phi(x)$. The imputed depth $d(x)$ is used to backproject the lifted coordinate $\bar x$ to yield a 3D positional encoding for each pixel $x_{3D} = K^{-1}\bar x d(x)$.

Here $\Omega \subset \real^2$ is discretized into a lattice of $H \times W$ pixels in the first layer, corresponding to the resolution of the original image, that decreases by a factor of 2 in each subsequent layer until the 5-th or last layer at $H/32 \times W/32$. Hence, the intrinsics parameters, focal lengths and principal point, must also be scaled by the same factor according to the resolution reduction in each layer.

The 3D positional encoding is concatenated with the image encoding $\psi(x) \in \real^N$, and, if available, the output of the previous KB layer $\psi_{3D}(x) \in \real^N$ where $N$ ranges from 48 in the first layer to 386 in the last. This is fused together by a $1 \times 1$ convolution to yield the output RGB 3D encoding. This encoding is fed to the next layer and also replaces the typical RGB skip connection to the decoder. Finally, the output depth and image encodings of the KB layer are produced by convolving separate $3 \times 3$ kernels. After which, both are also passed to the next layer as input.

In addition to benefits of generalization (\tabref{tab:nyuv2_test_set_results}), KB layers also produce depth estimates that better respect object boundaries. Because each layer encodes ``closeness'' based on the scene topology via 3D positional encoding rather than the 2D image topology (as in previous works), adjacent pixels in the image that are often confused to be close are now well separated (\figref{fig:kitti_test_set_head_to_head}) and hence distinct adjacent objects are better delineated and points belonging to the same surface are better regularized. This reduces the common bleed effect observed when a depth map is backprojected to a point cloud in 3D. 
Moreover, by instilling 3D structure as an architectural inductive bias, we enable a faster and slimmer network with fewer layers and parameters to achieve better performance (see \tabref{tab:kitti_test_set_results}, \ref{tab:void_test_set_results}). 

We note that our S2D module complements our KB layers as it provides us with dense or quasi-dense depth representation. Without it, we are left with sparse geometry, which limits the potential performance gain. Yet, as demonstrated in \tabref{tab:kitti_validation_set_results}, there are still benefits to using calibrated backprojection with a sparse representation.

\subsection{Loss Function}
\label{sec:loss_function}
Similar to previous works \cite{ma2019self,wong2021learning, wong2020unsupervised}, our loss function is the linear combination of three terms:
\begin{equation}
    \label{eqn:objective_function}
    \mathcal{L} = w_{ph}\ell_{ph}+w_{sz}\ell_{sz}+w_{sm}\ell_{sm}
\end{equation}
where $\ell_{ph}$ denotes photometric consistency, $\ell_{sz}$ sparse depth consistency, and $\ell_{sm}$ local smoothness. Each term is weighted by their associated $w$ (see \secref{sec:implementation_details}).

\textbf{Photometric Consistency.}
As mentioned in \secref{sec:network_architecture}, unsupervised methods leverage photometric reprojection error as a supervisory signal by reconstructing $I_{t}$ from $I_{\tau}$ for $\tau \in T \doteq \{t-1, t+1\}$ via \eqnref{eqn:image_reconstruction}. To accomplish this, one can obtain pose from a VIO \cite{fei2019geo} or employ a pose network to estimate the relative pose between $I_{t}$ and $I_{\tau}$ (see full system diagram in Supp. Mat.). We note that pose is only needed for training and is not used at test time.

From the reconstructions, the photometric consistency loss measures the average photometric reprojection error using a combination of $L^1$ penalty and \texttt{SSIM} \cite{wang2004image}:
\begin{equation}
\begin{aligned}
  	\ell_{ph} = \frac{1}{|\Omega|} 
  	    \sum_{\tau\in T} \sum_{x \in \Omega}  
  	        &w_{co}| \hat{I}_{\tau}(x)-I_{t}(x)| + \\ 
  	        &w_{st}\big(1 - \texttt{SSIM}(\hat{I}_{\tau}(x), I_t(x))\big),
\label{eqn:photometric_consistency_loss}
\end{aligned}
\end{equation}
$w_{co}$ and $w_{st}$ are weights for each term and are discussed in \secref{sec:implementation_details}. We note that if $g_{\tau t}$ is estimated via a pose network, instead of a VIO, it can be jointly learned with KBNet (\figref{fig:network_architecture}) as a by product from minimizing \eqnref{eqn:photometric_consistency_loss} and \ref{eqn:sparse_depth_consistency_loss}, and hence does not require any extra supervision. 

\begin{table}[t]
\centering
\footnotesize
\setlength\tabcolsep{18pt}
\begin{tabular}{l l}
    \midrule
        Metric & Definition \\ \midrule
        MAE &$\frac{1}{|\Omega|} \sum_{x\in\Omega} |\hat d(x) - d_{gt}(x)|$ \\
        RMSE & $\big(\frac{1}{|\Omega|}\sum_{x\in\Omega}|\hat d(x) - d_{gt}(x)|^2 \big)^{1/2}$ \\
        iMAE & $\frac{1}{|\Omega|} \sum_{x\in\Omega} |1/ \hat d(x) - 1/d_{gt}(x)|$ \\
        iRMSE& $\big(\frac{1}{|\Omega|}\sum_{x\in\Omega}|1 / \hat d(x) - 1/d_{gt}(x)|^2\big)^{1/2}$ \\ \midrule
    \end{tabular}
    \vspace{-0.8em}
    \caption{
        \textit{Error metrics.} $d_{gt}$ denotes the ground-truth depth.
    }
    \vspace{-1.6em}
\label{tab:error_metrics}
\end{table}

\textbf{Sparse Depth Consistency.}
Minimizing the reprojection error will reconstruct the scene structure up to an unknown scale. To ground the predictions to \textit{metric} scale, we minimize the $L^1$ difference between our predictions $\hat{d}$ and the sparse depth inputs over its domain ($\Omega_{z}$):
\begin{equation}
\label{eqn:sparse_depth_consistency_loss}
  	\ell_{sz} = \frac{1}{|\Omega_z|} 
  	    \sum_{x \in \Omega_z} 
  	        | \hat{d}(x) - z(x)|. 
\end{equation}

\textbf{Local Smoothness.}
We enforce local smoothness and connectivity over $\hat{d}$ by minimizing the $L^1$ penalty on its gradients in the $x-$ ($\partial_{X}$) and $y-$ ($\partial_{X}$) directions. We also weight each term using its respective image gradients, $\lambda_{X} = e^{-|\partial_{X}I_{t}(x)|}$ and $\lambda_{Y} = e^{-|\partial_{Y}I_{t}(x)|}$, to allow discontinuities along object boundaries:
\begin{equation}
\label{eqn:local_smoothness_loss}
  	\ell_{sm} = \frac{1}{|\Omega|}
  	    \sum_{x \in \Omega} 
      	    \lambda_{X}(x)|\partial_{X}\hat{d}(x)|+
      	    \lambda_{Y}(x)|\partial_{Y}\hat{d}(x)|.
\end{equation}

\section{Experiments and Results}
\label{sec:experiments_and_results}
We evaluate our method on benchmark datasets, KITTI \cite{uhrig2017sparsity} for outdoors settings, and VOID \cite{wong2020unsupervised} for indoors, using metrics describes in \tabref{tab:error_metrics}. We also demonstrate that our approach generalizes well to scenes captures by camera setup different than that used to collect the training set by training our model on VOID and testing it on NYUv2 \cite{silberman2012indoor}.

\vspace{-0.1em}
\subsection{Implementation Details}
\vspace{-0.1em}
\label{sec:implementation_details}
We implemented our method in PyTorch \cite{paszke2019pytorch}. End-to-end inference takes 16ms per frame. We used Adam \cite{kingma2015adam} with $\beta_1=0.9$ and $\beta_2=0.999$ to optimize our network. Training on KITTI \cite{uhrig2017sparsity} takes 70 hours for 60 epochs, VOID \cite{wong2020unsupervised} 16 hours for 15 epochs, and NYUv2 \cite{silberman2012indoor} 13 hours for 15 epochs on an Nvidia GTX 1080Ti GPU.  We use a batch size of 8 with $768 \times 320$ crops for KITTI, $640 \times 480$ for VOID and $576 \times 416$ for NYUv2. For KITTI, we choose $w_{ph} = 1$, $w_{co}=0.15$, $w_{st}=0.95$, $w_{sz} = 0.6$, and $w_{sm} = 0.04$; for VOID and NYUv2, we set $w_{sz} = 2$ and $w_{sm} = 2$. For detailed learning rate schedule, augmentations and S2D kernel sizes used for each dataset, please see Supp. Mat. 

\begin{figure*}[ht]
    \centering
    \includegraphics[width=1.0\linewidth]{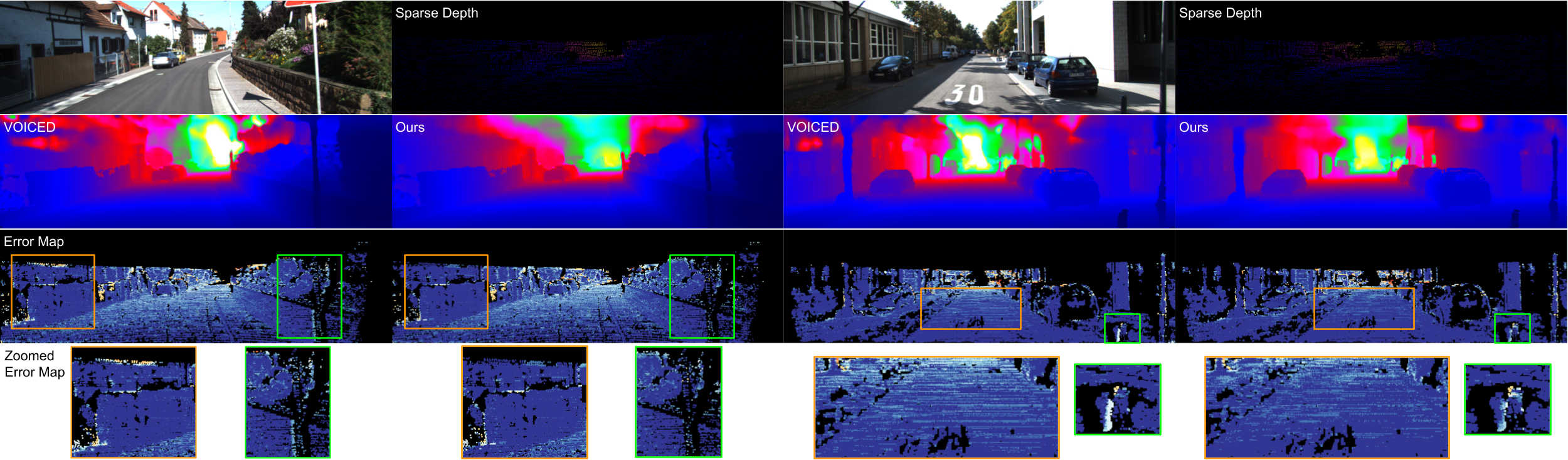}
    \caption{\textit{Qualitative results on KITTI test set.} Head-to-head comparison against \cite{wong2020unsupervised}. Thanks to our 3D positional encoding, our method performs well on regions where adjacent structures in 2D image space are far apart in the 3D scene e.g. street sign and wall (left panel, highlighted in green) and far region of the road (right panel, in orange).}
    \label{fig:kitti_test_set_head_to_head}
    \vspace{-1.3em}
\end{figure*}

\vspace{-0.1em}
\subsection{Datasets}
\vspace{-0.1em}
\textbf{KITTI} \cite{uhrig2017sparsity} provides $\approx$80,000 raw image frames and associated sparse depth maps. The sparse depth maps are the raw output from the Velodyne lidar sensor, each with a density of $\approx$5\%. Ground-truth depth is obtained by accumulating 11 neighbouring raw lidar scans. Semi-dense depth is available for the lower $30\%$ of the image space. We use the official 1,000 samples for validation and test on 1,000 designated samples (evaluated on their online test server). 

\textbf{VOID} \cite{wong2020unsupervised} contains synchronized $640 \times 480$ RGB images and sparse depth maps of indoor (laboratories, classrooms) and outdoor (gardens) scenes. $\approx 1500$ sparse depth  points (covering $\approx0.5\%$ of the image) are the set of features tracked by XIVO \cite{fei2019geo}, a VIO system. The ground-truth depth maps are dense and are acquired by active stereo. The entire dataset contains 56 sequences with challenging motion. Of the 56 sequences, 48 sequences ($\approx 40,000$) are designated for training and 8 for testing. The testing set contains $800$ frames. We follow the evaluation protocol of \cite{wong2020unsupervised} and cap the depths between 0.2 and 5 meters.

\begin{table}[t]
    \footnotesize
    \centering
    \setlength\tabcolsep{3.5pt}
    \begin{tabular}{l c c c c c c}
        \midrule
        Method & \# Param & Time & MAE & RMSE & iMAE & iRMSE \\ 
        \midrule
        SS-S2D \cite{ma2019self}
        & 27.8M & 80ms & 350.32 & 1299.85 & 1.57 & 4.07 \\ 
        \midrule
        IP-Basic \cite{ku2018defense} 
        & 0 & 11ms & 302.60 & 1288.46 & 1.29 & 3.78 \\ 
        \midrule
        DFuseNet \cite{shivakumar2019dfusenet}
        & n/a & 80ms & 429.93 & 1206.66 & 1.79 & 3.62 \\ 
        \midrule
        DDP* \cite{yang2019dense}
        & 18.8M & 80ms & 343.46 & 1263.19 & 1.32 & 3.58 \\ 
        \midrule
        VOICED \cite{wong2020unsupervised}  
        & 9.7M & 44ms & 299.41 & 1169.97 & 1.20 & 3.56 \\ 
        \midrule
        AdaFrame \cite{wong2021adaptive}
        & 6.4M & 40ms & 291.62 & 1125.67 & 1.16 & 3.32 \\ 
        \midrule
        SynthProj* \cite{lopez2020project}
        & 2.6M & 60ms & 280.42 & 1095.26 & 1.19 & 3.53 \\
        \midrule
        ScaffNet* \cite{wong2021learning}  
        & 7.8M & 32ms & 280.76 & 1121.93 & 1.15 & 3.30 \\ 
        \midrule
        Ours
        & 6.9M & 16ms & \textbf{256.76} & \textbf{1069.47} & \textbf{1.02} & \textbf{2.95} \\
        \midrule
    \end{tabular}
    \vspace{-0.8em}
    \caption{
        \textit{Quantitative results on the KITTI test set}. Our method outperforms all unsupervised methods across all metrics on the KITTI leaderboard. Compared to the the baseline \cite{wong2020unsupervised}, we improve by an average of 13.7\% across all metrics while using 29\% fewer parameters. * denotes methods that use additional synthetic data for training.
    }
\vspace{-1.3em}
\label{tab:kitti_test_set_results}
\end{table}

\textbf{NYUv2} \cite{silberman2012indoor} consists of 372K synchronized $640 \times 480$ RGB images and depth maps for 464 indoors scenes (household, offices, commercial), captured with a Microsoft Kinect. The official split consisting in 249 training and 215 test scenes. For training, we evenly sample a subset of the training split to yield 46K frames. We use the official validation set of 795 images and test set of 654 images. Because there are no sparse depth maps provided, we sampled  $\approx 1500$ points from the depth map via Harris corner detector \cite{harris1988combined} to mimic the sparse depth produced by SLAM/VIO. 

\begin{table}[t]
    \centering
    \footnotesize
    \setlength\tabcolsep{4.5pt}
    \begin{tabular}{l c c c c}
        \midrule 
        Method & MAE & RMSE & iMAE & iRMSE \\
        \midrule
        VOICED \cite{wong2020unsupervised} w/o Scaffolding
        & 347.14 & 1330.88 & 1.46 & 4.22 \\
        \midrule
        VOICED \cite{wong2020unsupervised} 
        & 305.06 & 1239.06 & 1.21 & 3.71 \\
        \midrule
        Ours w/o S2D
        & 287.76 & 1184.24 & 1.12 & 3.48 \\
        \midrule
        Ours w/o KB layers
        & 285.97 & 1171.88 & 1.11 & 3.40 \\
        \midrule
        Ours w/ Scaffolding \cite{wong2020unsupervised}
        & 275.56 & 1183.57 & 1.08 & 3.39 \\
        \midrule
        Ours w/ SPP \cite{he2015spatial,wong2021learning}
        & 273.08 & 1177.69 & 1.07 & 3.35 \\
        \midrule
        Ours
        & \textbf{260.44} & \textbf{1126.85} & \textbf{1.03} & \textbf{3.20} \\
        \midrule
    \end{tabular}
    \vspace{-0.8em}
    \caption{
        \textit{Ablation study on KITTI validation set.} Without S2D (row 3), our performance degrade because our 3D positional features will only encode sparse geometry, but we still beat \cite{wong2020unsupervised} in rows 1, 2 (``w/o Scaffolding'' is \cite{wong2020unsupervised} with sparse representation). We observe similar degradation without KB layers (row 6, replaced with VGG block used by \cite{wong2020unsupervised}). Substituting our S2D with Scaffolding \cite{wong2020unsupervised} or SPP \cite{he2015spatial,wong2021learning} also hurts performance (rows 7, 8). 
    }
\vspace{-1.3em}
\label{tab:kitti_validation_set_results}
\end{table}

\vspace{-0.1em}
\subsection{KITTI Depth Completion Benchmark}
\vspace{-0.1em}
We compare our method against recent unsupervised depth completion methods on the KITTI test set in \tabref{tab:kitti_test_set_results} (results taken from online leaderboard). Compared to the baseline \cite{wong2020unsupervised}, we improve by an average of 13.7\% across metrics and by as much as 17.1\% in iRMSE while reducing model size by 29\%. Overall, we beat the best performing method \cite{wong2021learning} by an average of 8.8\% and up to 10.6\% on the iMAE metric with a 11.5\% reduction in model size. We note that top methods \cite{lopez2020project,wong2021learning} use \textit{additional} synthetic data for training; whereas, we do not. Also, for inference, our method takes 16ms per image (62 FPS), which is $2.75\times$ faster than  \cite{wong2020unsupervised}\footnote{The reported run time of \cite{wong2020unsupervised} on the KITTI leaderboard did not include their scaffolding step; whereas, the number in \tabref{tab:kitti_test_set_results} accounts for it.} and $2\times$ faster than the state of the art \cite{wong2021learning}. We note that our method significantly improves the iMAE and iRMSE metrics, to the point where we are comparable to some of the supervised methods for close range performance. For example, our iMAE score is ranked 5th across all methods (see \tabref{tab:kitti_supervised_benchmark}, \ref{tab:kitti_unsupervised_benchmark} in Supp. Mat.). To the best of our knowledge, we are the first work in unsupervised depth completion to demonstrate comparable performance to supervised methods. 

To show the improvements from our contributions, we show head-to-head qualitative comparisons against the baseline \cite{wong2020unsupervised} in \figref{fig:kitti_test_set_head_to_head}. Our method performs better in regions where depth discontinuities occur in image topology i.e. street sign and wall (left panel, highlighted in green) and far regions of the road (right panel, in orange). This is in part thanks to our calibrated backprojection (KB) layer which goes counter to the current trend of learning everything with generic architectures, including what we already know about basic Euclidean geometry. Our KB layers imposes strong inductive bias by incorporating the camera intrinsic calibration matrix to yield 3D positional encoding that lifts the image representation into scene topology -- this delineates points where in 2D image topology are ``close'', but can be far in 3D scene topology.

\begin{figure}[t]
    \centering
    \includegraphics[width=0.8\linewidth]{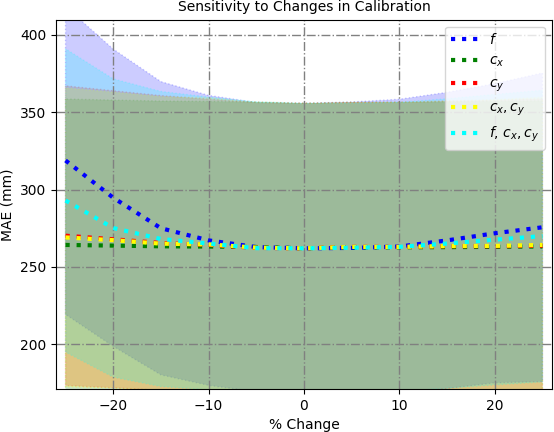}
    \caption{
        \textit{Sensitivity to changes in calibration on KITTI.} Focal length and principal point are altered to test sensitivity to changes in intrinsics parameters. Our method is robust to change up to $\approx10\%$. After which, performance degrades. 
    } 
\vspace{-1.3em}
\label{fig:kitti_validation_set_calibration_sensitivity}
\end{figure}

\tabref{tab:kitti_validation_set_results} shows an ablation study on the KITTI validation set. As mentioned in \secref{sec:network_architecture}, our sparse-to-dense module (S2D) provides dense depth representation which in turn enables dense 3D topology in our calibrated backprojection (KB) layers. Hence, removing it (``w/o S2D) will hurt performance because it results in a \textit{sparse} 3D positional encoding. Nonetheless, sparse geometry is still helpful as we outperform \cite{wong2020unsupervised} in rows 1, 2. Similarly, replacing our KB (``w/o KB layers'') with VGG blocks used by \cite{wong2020unsupervised} also hurts performance as the model now lacks 3D spatial position. We show in rows 5 and 6 that one cannot simply substitute S2D with scaffolding \cite{wong2020unsupervised} or SPP \cite{he2015spatial,wong2021learning}. 

In \figref{fig:kitti_validation_set_calibration_sensitivity}, we perform a  sensitivity study of our model to calibration on the KITTI validation set. To this end, we altered the calibration by increasing or decreasing focal length ($f$) and/or principal point ($c_x, c_y$) and feed it as input. Our model is robust to changes up to $\approx10\%$; after which, performance degrades. While changes in $c_x, c_y$ have minor effects (which is scene-dependent), we observe a sharp decrease in performance when we decrease $f$ by 20 to 25\%. This is because, geometrically, decreasing $f$ backprojects points to a larger field of view, distorting surfaces and sending points of the same surface far from each other. Increasing  $f$ conversely ``packs'' them tighter; this is okay for small increases, but for larger values, points will get ``squashed together'' -- thus hurting performance. Also, to quantify the effect of sparsity, we provide a sensitivity study on various density levels in Supp. Mat.

\begin{figure}[t]
    \centering
    \includegraphics[width=0.95\linewidth]{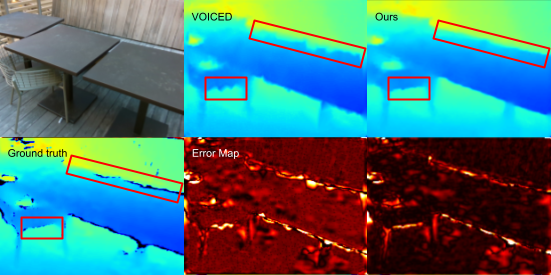}
    \vspace{-0.5em}
    \caption{
        \textit{Qualitative results on VOID test set.} Comparison against \cite{wong2020unsupervised}. Our method performs better overall. 
    } 
\vspace{-0.8em}
\label{fig:void_test_set}
\end{figure}

\begin{table}[t]
    \footnotesize
    \centering
    \setlength\tabcolsep{3.5pt}
    \begin{tabular}{l c c c c c c}
        \midrule 
        Method & \# Param & Time & MAE & RMSE & iMAE & iRMSE \\ 
        \midrule
        SS-S2D \cite{ma2019self} 
        & 27.8M & 59ms & 178.85 & 243.84 & 80.12 & 107.69 \\ 
        \midrule 
        DDP \cite{yang2019dense} 
        & 18.8M & 54ms & 151.86 & 222.36 & 74.59 & 112.36 \\ 
        \midrule 
        VOICED \cite{wong2020unsupervised} 
        & 9.7M & 29ms & 85.05 & 169.79 & 48.92 & 104.02 \\ 
        \midrule
        ScaffNet \cite{wong2021learning}
        & 7.8M & 25ms & 59.53 & 119.14 & 35.72 & 68.36 \\ 
        \midrule
        Ours
        & 6.9M & 13ms & \textbf{39.80} & \textbf{95.86} & \textbf{21.16} & \textbf{49.72} \\
        \midrule
    \end{tabular}
    \vspace{-0.8em}
    \caption{\textit{Quantitative results on VOID test set.} 
        We outperform all competing methods across all metrics. Compared to \cite{wong2021learning}, we improve by an average of 30.5\%.
    }
\vspace{-1.4em}
\label{tab:void_test_set_results}
\end{table}

\begin{table}[t]
    \footnotesize
    \centering
    \setlength\tabcolsep{5pt}
    \begin{tabular}{l c c c c c}
        \midrule 
        Method & Trained on & MAE & RMSE & iMAE & iRMSE \\ 
        \midrule
        VOICED \cite{wong2020unsupervised} 
        & NYUv2 & 127.61 & 228.38 & 28.89 & 54.70 \\ 
        \midrule
        VOICED \cite{wong2020unsupervised} 
        & VOID & 178.87 & 329.28 & 42.57 & 105.93 \\ 
        \midrule
        ScaffNet \cite{wong2021learning} 
        & NYUv2 & 117.49 & 199.31 & 24.89 & 44.06 \\ 
        \midrule
        ScaffNet \cite{wong2021learning} 
        & VOID & 155.20 & 241.42 & 31.77 & 52.62 \\
        \midrule
        Ours 
        & NYUv2 & 105.76 & 197.77 & 21.37 & 42.74 \\ 
        \midrule
        Ours 
        & VOID & 117.18 & 218.67 & 23.01 &	47.96 \\ 
        \midrule
    \end{tabular}
    \vspace{-0.8em}
    \caption{\textit{Quantitative results on the NYUv2 test set.} Column titled ``Trained on'' denotes the dataset each method is trained on. \cite{wong2021learning,wong2020unsupervised} degrade much more than our method when tested on a dataset captured by a different sensor platform than the one used for gathering its training data.}
\vspace{-1.4em}
\label{tab:nyuv2_test_set_results}
\end{table}

\vspace{-0.2em}
\subsection{VOID Depth Completion Benchmark}
\vspace{-0.2em}
In the indoor scenario, the point clouds are on orders of hundreds to several thousand points (if we are being generous); hence, because of the sparsity, perturbations to the point cloud can yield vastly different sparse geometry. This increases a model's sensitivity to the distribution to the sparse points. As there exists many complex scene layouts for the indoor setting, learning a dense representation and understanding the 3D topology of the scene become even more important. This is shown in \tabref{tab:void_test_set_results} where we outperform \cite{ma2019self,wong2021learning,wong2020unsupervised,yang2019dense} across all metrics to achieve the state of the art on VOID. A key comparison is between our method and \cite{wong2020unsupervised}. Even though \cite{wong2020unsupervised} creates a hand-crafted scaffolding of the scene to obtain a dense representation, because there are very few points, it is prone to error i.e. forming surfaces between discontinuous objects and sensitive to changes in the points sampled. This is where our method shines. By optimizing for the trade-off between density and detail, our S2D module learns to exploit the natural statistics of the dataset to obtain a dense representation more compatible with the scene. Also, our KB layers introduces 3D topology as an inductive bias, allowing the network to delineate points that are close in image topology, but are far in scene topology -- culminating in 51.7\% and 30.5\% improvement over \cite{wong2020unsupervised} and the state of the art \cite{wong2021learning}, respectively. 

In \tabref{tab:nyuv2_test_set_results}, we show that our method generalizes well to sensor platforms not used in the training set by training our method on VOID (captured on Intel RealSense) and testing it on NYUv2 (Microsoft Kinect). Similarly, we test models pretrained on VOID released by \cite{wong2021learning,wong2020unsupervised} on NYUv2. We also train our method and \cite{wong2021learning,wong2020unsupervised} from scratch on NYUv2 to show the paragon performance (rows 1, 3, 5). Rows 1 shows that \cite{wong2020unsupervised} does not generalize well to NYUv2 where error increases by 56\% (as much as 94\% in iRMSE). While \cite{wong2021learning} does better, there is still a sharp decrease of 25.1\% in performance. This is in part due to the change in sensor platform as well scene distribution in NYUv2. While we do not achieve paragon performance, our method generalizes better with a reasonable 10\% increase in error -- improving over \cite{wong2020unsupervised} by 83\% and \cite{wong2021learning} by 62\% in relative error. We note that while training on the full set for NYUv2 should yield better results for paragon performance, our model trained on VOID performs better than VOICED \cite{wong2020unsupervised} and comparable to ScaffNet \cite{wong2020unsupervised} trained on the subset of NYUv2. For qualitative comparisons, please see \figref{fig:void_to_nyuv2_head_to_head} in Supp. Mat.

\vspace{-0.4em}
\section{Discussion}
\vspace{-0.3em}
We present an approach to unsupervised depth completion that imposes strong inductive biases on Euclidean reconstruction in the architecture, rather than learning from data with a generic model such as a Transformer. This presents some advantages. First, it allows feeding calibration as an input, which means that we can easily use a model trained with a certain sensor platform with a different one at inference time. Second, the calibrated backprojection layer explicitly incorporates a basic geometric image formation model based on Euclidean transformations in 3D and central perspective projection onto 2D. This allows us to reduce the model size while still achieving the state of the art.  

However, imposing strong inductive biases also presents some risks and limitations. First, if the camera is miscalibrated, inputing the wrong calibration can backfire, yielding distorted depth maps. Second, only a very rudimentary calibration model is used, so if a sensor platform has fancy optics such as omnidirectional lenses, one cannot use one of our pre-trained models but rather has to modify the core backprojection module. Third, even with these ad-hoc architectural choices, our model suffers the limitations of all imputations, which is that where there is insufficient evidence to constrain the solution, the regularizer dominates, which is a form of hallucination and can yield wildly wrong inferences. This would be mitigated by having an accurate measure of uncertainty associated to the depth map, this is an open problem well beyond our focus here.

\textbf{Acknowledgements.} This work was supported by ARL W911NF-20-1-0158 and ONR N00014-17-1-2072.

{\small
\bibliographystyle{ieee_fullname}
\bibliography{egbib}

\begin{thebibliography}{10}\itemsep=-1pt

\bibitem{blomgren1998color}
Peter Blomgren and Tony~F Chan.
\newblock Color tv: total variation methods for restoration of vector-valued
  images.
\newblock {\em IEEE transactions on image processing}, 7(3):304--309, 1998.

\bibitem{chang2018pyramid}
Jia-Ren Chang and Yong-Sheng Chen.
\newblock Pyramid stereo matching network.
\newblock In {\em Proceedings of the IEEE Conference on Computer Vision and
  Pattern Recognition}, pages 5410--5418, 2018.

\bibitem{chen2019learning}
Yun Chen, Bin Yang, Ming Liang, and Raquel Urtasun.
\newblock Learning joint 2d-3d representations for depth completion.
\newblock In {\em Proceedings of the IEEE International Conference on Computer
  Vision}, pages 10023--10032, 2019.

\bibitem{cheng2020cspn++}
Xinjing Cheng, Peng Wang, Chenye Guan, and Ruigang Yang.
\newblock Cspn++: Learning context and resource aware convolutional spatial
  propagation networks for depth completion.
\newblock In {\em Proceedings of the AAAI Conference on Artificial
  Intelligence}, volume~34, pages 10615--10622, 2020.

\bibitem{cheng2018depth}
Xinjing Cheng, Peng Wang, and Ruigang Yang.
\newblock Depth estimation via affinity learned with convolutional spatial
  propagation network.
\newblock In {\em Proceedings of the European Conference on Computer Vision
  (ECCV)}, pages 103--119, 2018.

\bibitem{chodosh2018deep}
Nathaniel Chodosh, Chaoyang Wang, and Simon Lucey.
\newblock Deep convolutional compressed sensing for lidar depth completion.
\newblock In {\em Asian Conference on Computer Vision}, pages 499--513.
  Springer, 2018.

\bibitem{dimitrievski2018learning}
Martin Dimitrievski, Peter Veelaert, and Wilfried Philips.
\newblock Learning morphological operators for depth completion.
\newblock In {\em International Conference on Advanced Concepts for Intelligent
  Vision Systems}. Springer, 2018.

\bibitem{eldesokey2020uncertainty}
Abdelrahman Eldesokey, Michael Felsberg, Karl Holmquist, and Michael Persson.
\newblock Uncertainty-aware cnns for depth completion: Uncertainty from
  beginning to end.
\newblock In {\em Proceedings of the IEEE/CVF Conference on Computer Vision and
  Pattern Recognition}, pages 12014--12023, 2020.

\bibitem{eldesokey2018propagating}
Abdelrahman Eldesokey, Michael Felsberg, and Fahad~Shahbaz Khan.
\newblock Propagating confidences through cnns for sparse data regression.
\newblock In {\em Proceedings of British Machine Vision Conference (BMVC)},
  2018.

\bibitem{facil2019cam}
Jose~M Facil, Benjamin Ummenhofer, Huizhong Zhou, Luis Montesano, Thomas Brox,
  and Javier Civera.
\newblock Cam-convs: Camera-aware multi-scale convolutions for single-view
  depth.
\newblock In {\em Proceedings of the IEEE/CVF Conference on Computer Vision and
  Pattern Recognition}, 2019.

\bibitem{fei2019geo}
Xiaohan Fei, Alex Wong, and Stefano Soatto.
\newblock Geo-supervised visual depth prediction.
\newblock {\em IEEE Robotics and Automation Letters}, 4(2):1661--1668, 2019.

\bibitem{fischler1981random}
Martin~A Fischler and Robert~C Bolles.
\newblock Random sample consensus: a paradigm for model fitting with
  applications to image analysis and automated cartography.
\newblock {\em Communications of the ACM}, 24(6):381--395, 1981.

\bibitem{gaidon2016virtual}
Adrien Gaidon, Qiao Wang, Yohann Cabon, and Eleonora Vig.
\newblock Virtual worlds as proxy for multi-object tracking analysis.
\newblock In {\em Proceedings of the IEEE conference on computer vision and
  pattern recognition}, 2016.

\bibitem{godard2019digging}
Cl{\'e}ment Godard, Oisin Mac~Aodha, Michael Firman, and Gabriel~J Brostow.
\newblock Digging into self-supervised monocular depth estimation.
\newblock In {\em Proceedings of the IEEE/CVF International Conference on
  Computer Vision}, 2019.

\bibitem{gordon2019depth}
Ariel Gordon, Hanhan Li, Rico Jonschkowski, and Anelia Angelova.
\newblock Depth from videos in the wild: Unsupervised monocular depth learning
  from unknown cameras.
\newblock In {\em Proceedings of the IEEE/CVF International Conference on
  Computer Vision}, pages 8977--8986, 2019.

\bibitem{handa2016gvnn}
Ankur Handa, Michael Bloesch, Viorica P{\u{a}}tr{\u{a}}ucean, Simon Stent, John
  McCormac, and Andrew Davison.
\newblock gvnn: Neural network library for geometric computer vision.
\newblock In {\em European Conference on Computer Vision}. Springer, 2016.

\bibitem{harris1988combined}
Christopher~G Harris, Mike Stephens, et~al.
\newblock A combined corner and edge detector.
\newblock In {\em Alvey vision conference}, volume~15, pages 10--5244.
  Citeseer, 1988.

\bibitem{he2015spatial}
Kaiming He, Xiangyu Zhang, Shaoqing Ren, and Jian Sun.
\newblock Spatial pyramid pooling in deep convolutional networks for visual
  recognition.
\newblock {\em IEEE transactions on pattern analysis and machine intelligence},
  37(9):1904--1916, 2015.

\bibitem{huang2019hms}
Zixuan Huang, Junming Fan, Shenggan Cheng, Shuai Yi, Xiaogang Wang, and
  Hongsheng Li.
\newblock Hms-net: Hierarchical multi-scale sparsity-invariant network for
  sparse depth completion.
\newblock {\em IEEE Transactions on Image Processing}, 29:3429--3441, 2019.

\bibitem{jaritz2018sparse}
Maximilian Jaritz, Raoul De~Charette, Emilie Wirbel, Xavier Perrotton, and
  Fawzi Nashashibi.
\newblock Sparse and dense data with cnns: Depth completion and semantic
  segmentation.
\newblock In {\em 2018 International Conference on 3D Vision (3DV)}, pages
  52--60. IEEE, 2018.

\bibitem{kingma2015adam}
Diederik~P Kingma and Jimmy~Lei Ba.
\newblock Adam: A method for stochastic gradient descent.
\newblock In {\em ICLR: International Conference on Learning Representations},
  2015.

\bibitem{ku2018defense}
Jason Ku, Ali Harakeh, and Steven~L Waslander.
\newblock In defense of classical image processing: Fast depth completion on
  the cpu.
\newblock In {\em 2018 15th Conference on Computer and Robot Vision (CRV)},
  pages 16--22. IEEE, 2018.

\bibitem{lepetit2009epnp}
Vincent Lepetit, Francesc Moreno-Noguer, and Pascal Fua.
\newblock Epnp: An accurate o (n) solution to the pnp problem.
\newblock {\em International journal of computer vision}, 81(2):155, 2009.

\bibitem{li2020multi}
Ang Li, Zejian Yuan, Yonggen Ling, Wanchao Chi, Chong Zhang, et~al.
\newblock A multi-scale guided cascade hourglass network for depth completion.
\newblock In {\em Proceedings of the IEEE/CVF Winter Conference on Applications
  of Computer Vision}, pages 32--40, 2020.

\bibitem{lopez2020project}
Adrian Lopez-Rodriguez, Benjamin Busam, and Krystian Mikolajczyk.
\newblock Project to adapt: Domain adaptation for depth completion from noisy
  and sparse sensor data.
\newblock In {\em Proceedings of the Asian Conference on Computer Vision},
  2020.

\bibitem{ma2019self}
Fangchang Ma, Guilherme~Venturelli Cavalheiro, and Sertac Karaman.
\newblock Self-supervised sparse-to-dense: Self-supervised depth completion
  from lidar and monocular camera.
\newblock In {\em International Conference on Robotics and Automation (ICRA)},
  pages 3288--3295. IEEE, 2019.

\bibitem{ma2012invitation}
Yi Ma, Stefano Soatto, Jana Kosecka, and S~Shankar Sastry.
\newblock {\em An invitation to 3-d vision: from images to geometric models},
  volume~26.
\newblock Springer Science \& Business Media, 2012.

\bibitem{merrill2021robust}
Nathaniel Merrill, Patrick Geneva, and Guoquan Huang.
\newblock Robust monocular visual-inertial depth completion for embedded
  systems.
\newblock In {\em International Conference on Robotics and Automation (ICRA)}.
  IEEE, 2021.

\bibitem{mumford1989optimal}
David~Bryant Mumford and Jayant Shah.
\newblock Optimal approximations by piecewise smooth functions and associated
  variational problems.
\newblock {\em Communications on pure and applied mathematics}, 1989.

\bibitem{park2020non}
Jinsun Park, Kyungdon Joo, Zhe Hu, Chi-Kuei Liu, and In-So Kweon.
\newblock Non-local spatial propagation network for depth completion.
\newblock In {\em European Conference on Computer Vision, ECCV 2020}. European
  Conference on Computer Vision, 2020.

\bibitem{paszke2019pytorch}
Adam Paszke, Sam Gross, Francisco Massa, Adam Lerer, James Bradbury, Gregory
  Chanan, Trevor Killeen, Zeming Lin, Natalia Gimelshein, Luca Antiga, et~al.
\newblock Pytorch: An imperative style, high-performance deep learning library.
\newblock {\em Advances in Neural Information Processing Systems},
  32:8026--8037, 2019.

\bibitem{qiu2019deeplidar}
Jiaxiong Qiu, Zhaopeng Cui, Yinda Zhang, Xingdi Zhang, Shuaicheng Liu, Bing
  Zeng, and Marc Pollefeys.
\newblock Deeplidar: Deep surface normal guided depth prediction for outdoor
  scene from sparse lidar data and single color image.
\newblock In {\em Proceedings of the IEEE Conference on Computer Vision and
  Pattern Recognition}, pages 3313--3322, 2019.

\bibitem{qu2021bayesian}
Chao Qu, Wenxin Liu, and Camillo~J Taylor.
\newblock Bayesian deep basis fitting for depth completion with uncertainty.
\newblock {\em arXiv preprint arXiv:2103.15254}, 2021.

\bibitem{qu2020depth}
Chao Qu, Ty Nguyen, and Camillo Taylor.
\newblock Depth completion via deep basis fitting.
\newblock In {\em Proceedings of the IEEE/CVF Winter Conference on Applications
  of Computer Vision}, pages 71--80, 2020.

\bibitem{riba2020kornia}
Edgar Riba, Dmytro Mishkin, Daniel Ponsa, Ethan Rublee, and Gary Bradski.
\newblock Kornia: an open source differentiable computer vision library for
  pytorch.
\newblock In {\em Proceedings of the IEEE/CVF Winter Conference on Applications
  of Computer Vision}, pages 3674--3683, 2020.

\bibitem{rudin1992nonlinear}
Leonid~I Rudin, Stanley Osher, and Emad Fatemi.
\newblock Nonlinear total variation based noise removal algorithms.
\newblock {\em Physica D: nonlinear phenomena}, 60(1-4):259--268, 1992.

\bibitem{sartipi2020deep}
Kourosh Sartipi, Tien Do, Tong Ke, Khiem Vuong, and Stergios~I Roumeliotis.
\newblock Deep depth estimation from visual-inertial slam.
\newblock In {\em 2020 IEEE/RSJ International Conference on Intelligent Robots
  and Systems (IROS)}. IEEE, 2020.

\bibitem{schneider2016semantically}
Nick Schneider, Lukas Schneider, Peter Pinggera, Uwe Franke, Marc Pollefeys,
  and Christoph Stiller.
\newblock Semantically guided depth upsampling.
\newblock In {\em German conference on pattern recognition}, pages 37--48.
  Springer, 2016.

\bibitem{shivakumar2019dfusenet}
Shreyas~S Shivakumar, Ty Nguyen, Ian~D Miller, Steven~W Chen, Vijay Kumar, and
  Camillo~J Taylor.
\newblock Dfusenet: Deep fusion of rgb and sparse depth information for image
  guided dense depth completion.
\newblock In {\em 2019 IEEE Intelligent Transportation Systems Conference
  (ITSC)}, pages 13--20. IEEE, 2019.

\bibitem{silberman2012indoor}
Nathan Silberman, Derek Hoiem, Pushmeet Kohli, and Rob Fergus.
\newblock Indoor segmentation and support inference from rgbd images.
\newblock In {\em European conference on computer vision}, pages 746--760.
  Springer, 2012.

\bibitem{uhrig2017sparsity}
Jonas Uhrig, Nick Schneider, Lukas Schneider, Uwe Franke, Thomas Brox, and
  Andreas Geiger.
\newblock Sparsity invariant cnns.
\newblock In {\em 2017 International Conference on 3D Vision (3DV)}, pages
  11--20. IEEE, 2017.

\bibitem{van2019sparse}
Wouter Van~Gansbeke, Davy Neven, Bert De~Brabandere, and Luc Van~Gool.
\newblock Sparse and noisy lidar completion with rgb guidance and uncertainty.
\newblock In {\em 2019 16th International Conference on Machine Vision
  Applications (MVA)}, pages 1--6. IEEE, 2019.

\bibitem{vaswani2017attention}
Ashish Vaswani, Noam Shazeer, Niki Parmar, Jakob Uszkoreit, Llion Jones,
  Aidan~N Gomez, Lukasz Kaiser, and Illia Polosukhin.
\newblock Attention is all you need.
\newblock {\em arXiv preprint arXiv:1706.03762}, 2017.

\bibitem{wang2004image}
Zhou Wang, Alan~C Bovik, Hamid~R Sheikh, and Eero~P Simoncelli.
\newblock Image quality assessment: from error visibility to structural
  similarity.
\newblock {\em IEEE transactions on image processing}, 13(4):600--612, 2004.

\bibitem{wong2021learning}
Alex Wong, Safa Cicek, and Stefano Soatto.
\newblock Learning topology from synthetic data for unsupervised depth
  completion.
\newblock {\em IEEE Robotics and Automation Letters}, 6(2):1495--1502, 2021.

\bibitem{wong2021adaptive}
Alex Wong, Xiaohan Fei, Byung-Woo Hong, and Stefano Soatto.
\newblock An adaptive framework for learning unsupervised depth completion.
\newblock {\em IEEE Robotics and Automation Letters}, 6(2):3120--3127, 2021.

\bibitem{wong2020unsupervised}
Alex Wong, Xiaohan Fei, Stephanie Tsuei, and Stefano Soatto.
\newblock Unsupervised depth completion from visual inertial odometry.
\newblock {\em IEEE Robotics and Automation Letters}, 2020.

\bibitem{wong2019bilateral}
Alex Wong and Stefano Soatto.
\newblock Bilateral cyclic constraint and adaptive regularization for
  unsupervised monocular depth prediction.
\newblock In {\em Proceedings of the IEEE/CVF Conference on Computer Vision and
  Pattern Recognition}, pages 5644--5653, 2019.

\bibitem{xu2019depth}
Yan Xu, Xinge Zhu, Jianping Shi, Guofeng Zhang, Hujun Bao, and Hongsheng Li.
\newblock Depth completion from sparse lidar data with depth-normal
  constraints.
\newblock In {\em Proceedings of the IEEE International Conference on Computer
  Vision}, pages 2811--2820, 2019.

\bibitem{yang2019dense}
Yanchao Yang, Alex Wong, and Stefano Soatto.
\newblock Dense depth posterior (ddp) from single image and sparse range.
\newblock In {\em Proceedings of the IEEE/CVF Conference on Computer Vision and
  Pattern Recognition}, pages 3353--3362, 2019.

\bibitem{zhang2018deep}
Yinda Zhang and Thomas Funkhouser.
\newblock Deep depth completion of a single rgb-d image.
\newblock In {\em Proceedings of the IEEE Conference on Computer Vision and
  Pattern Recognition}, pages 175--185, 2018.

\bibitem{zhang2000flexible}
Zhengyou Zhang.
\newblock A flexible new technique for camera calibration.
\newblock {\em IEEE Transactions on pattern analysis and machine intelligence},
  22(11):1330--1334, 2000.

\bibitem{zuo2021codevio}
Xingxing Zuo, Nathaniel Merrill, Wei Li, Yong Liu, Marc Pollefeys, and Guoquan
  Huang.
\newblock Codevio: Visual-inertial odometry with learned optimizable dense
  depth.
\newblock In {\em International Conference on Robotics and Automation (ICRA)}.
  IEEE, 2021.

\end{thebibliography}
}

\newpage

\twocolumn[{%
\renewcommand\twocolumn[1][]{#1}%
\maketitle
\begin{center}
    \centering
    \includegraphics[width=\textwidth]{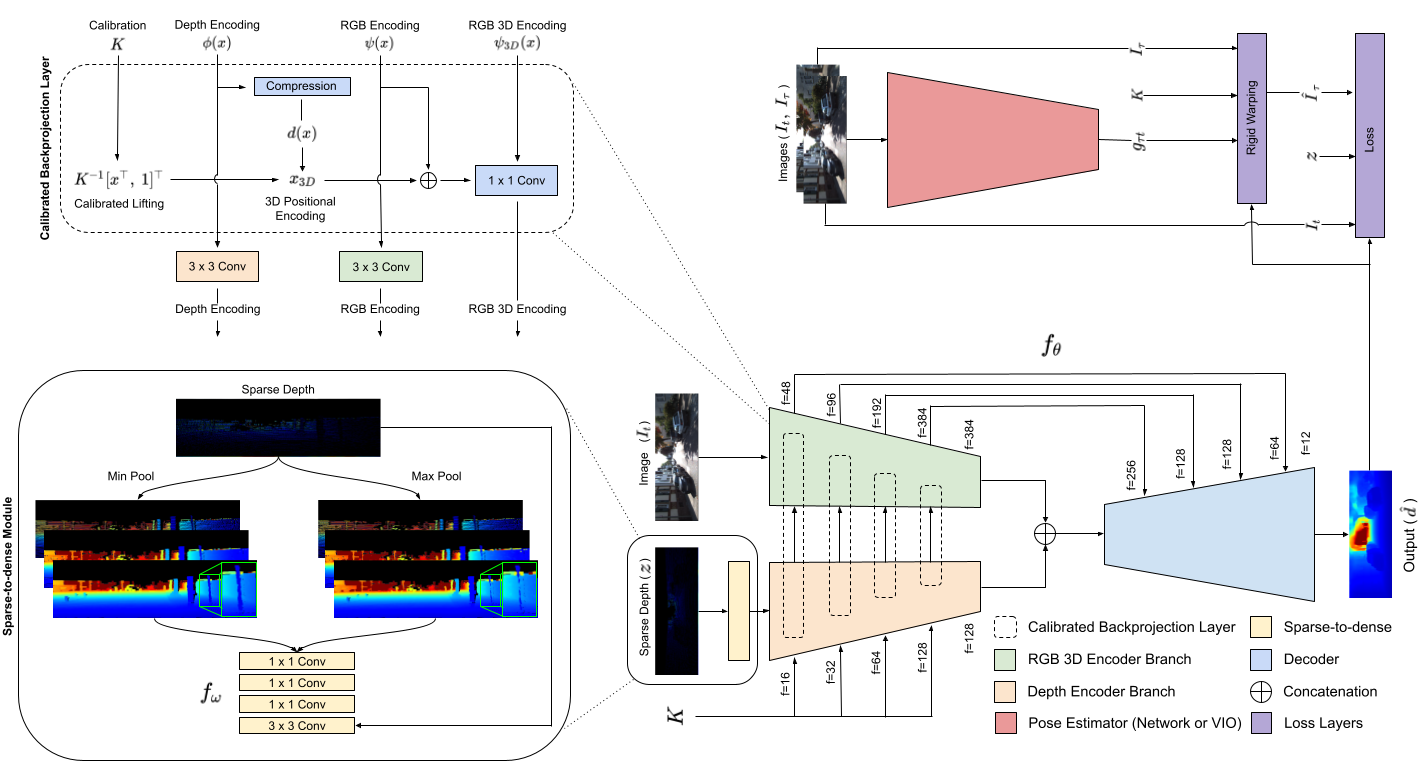}
    \captionof{figure}{\textit{System diagram during training}. We assume we are given monocular video sequences, synchronized sparse point clouds projected onto the image plane as 2.5D depth maps, and camera calibration. A training sample is therefore $(I_t, I_\tau, z, K)$. Sparse depth inputs ($z$) are fed to our sparse-to-dense module ($f_\omega$) to yield a dense or quasi-dense representation. Along with image ($I_t$) and camera calibration matrix ($K$), it is then fed into our depth completion network ($f_\theta$) comprised of calibrated backprojection layers to produce dense depth prediction $\hat{d}$. Relative pose ($g_{\tau t}$) between images $I_t$ and $I_\tau$ can be estimated from a VIO or a network. In the case of the latter, pose can be jointly learned with depth. We note that pose is only needed to give the reconstruction $\hat I_\tau$ for constructing the loss function and is not needed during inference. \\ \\
    Code available at: \url{ https://github.com/alexklwong/calibrated-backprojection-network}.}
    \vspace{1.5em}
\label{fig:system_diagram}
\end{center}
}]
\appendix

\begin{center} 
    {\LARGE{\textbf{ \\ Supplementary Materials}}}
\end{center}

\vspace{0.5em}

\noindent \textbf{Summary of contents.}
In \secref{sec:system_overview}, we provide an overview of our full system and more details on our loss function. We also provide the kernel sizes used in our sparse-to-dense module, augmentations used during training and our learning rate schedule to reproduce our results on KITTI \cite{uhrig2017sparsity}, VOID \cite{wong2020unsupervised}, and NYUv2 \cite{silberman2012indoor}. In \secref{sec:features_learned_sparse_to_dense}, we visualize and compare features learned by our proposed sparse-to-dense module to those from typical convolutional block, and show that our spare-to-dense module yields a much denser representation for the the depth completion network to ingest. In \secref{sec:sensitivity_incorrect_calibration}, we consider the possibility of miscalibration and examine the sensitivity of our model to changes in intrinsics parameters i.e. \textit{incorrect} calibration. We show that our model is robust to reasonable ranges of calibration error. In \secref{sec:sensitivity_density_levels}, we study the sensitivity of our model to changes in sparse depth input density levels and demonstrate that we are robust even when sparse point cover only 0.15\% of the image space. In \secref{sec:generalization_to_other_sensor_platforms}, we discuss our method's ability to generalize to test time sensor platforms with a different camera than the one used in training. Finally, in \secref{sec:kitti_depth_completion_benchmark}, we show that we can beat several supervised methods on KITTI online leaderboard and that we rank 5th amongst \textit{all methods} for the iMAE metric.

\section{System Overview}
\label{sec:system_overview}
\figref{fig:system_diagram} shows a diagram of our full system. Our model takes an RGB image $I$, a sparse depth map $z$, and the camera intrinsics matrix $K$ as input. First, the sparse depth map $z$ is fed into our sparse-to-dense module $f_\omega$ to obtain a dense or qusai-dense representation (Sec. 2.1, main text). Then, the depth representation $f_\omega(z)$, RGB image $I$, and intrinsics $K$ are fed into the depth completion network $f_\theta$, which is comprised of an encoder with calibrated backprojection layer followed by a decoder (Sec. 2.2, main text). Each calibrated backprojection realizes the backprojection process into 3D camera space by performing calibrated lifting of pixel coordinates using $K$, and projecting the depth representation to 1 dimension and multiplying it with the lifted coordinates -- result of which is a 3D positional encoding of the scene structure. 

To yield a unified depth and RGB representation, the 3D positional encoding from the depth branch is passed laterally to the RGB branch to enable association between each RGB feature and its 3D position. By doing so, we introduce 3D structure as an architectural inductive bias, which allows the network to reason about ``close'' points in the 2D image topology that are actually far in 3D scene topology. The RGB 3D representation is finally fed through the decoder to produce the final depth prediction $\hat d$.

\subsection{Loss Function}
To train our model, we assume the availability of previous and next RGB frames $I_\tau$ of the given image $I$ or $I_t$ (to denote the current time frame) where $\tau \in T \doteq \{t-1, t+1\}$. During training, we estimate the relative pose $g_{\tau t}$ between images at time $t$ and $\tau$. Using $I_\tau$, $K$ and $g_{\tau t}$, we can create the reconstruction $\hat{I}_t$ of $I_t$ via reprojection (Eqn. 4, main text) to enable an unsupervised loss (Eqn. 6-9, main text), which include a photometric reconstruction term, a sparse depth reconstruction term and a local smoothness term. 

We note that the photometric term can be replaced with more sophisticated measures of reprojection error \cite{godard2019digging} and additional regularizers such as pose consistency \cite{wong2020unsupervised} or adaptive regularization weighting schemes \cite{wong2019bilateral,wong2021adaptive} -- which would likely boost performance even more. However, we choose a simple loss to demonstrate the efficacy of our novel architecture. We note that $g_{\tau t}$ can be obtained by the means of a visual inertial odometry (VIO) system or a pose network if the VIO is not available. In the case where pose is obtained from network, the pose network can be trained jointly with our depth completion network (KBNet). Relative pose is learned as a byproduct of minimizing Eqn. 6 in main text. Also, since $g_{\tau t}$ is only need for reprojection during training; hence, the VIO system and the pose network are not necessary for inference. Because our network is fast  and light-weight (16ms run time per image, 6.9M parameters and 2.6GB memory as benchmarked on $1216 \times 352$ images from KITTI \cite{uhrig2017sparsity}), it can be deployed with a VIO system to learn online.

\begin{table}[t]
    \centering
    \footnotesize
    \setlength\tabcolsep{40pt}
    \begin{tabular}{l l}
        \midrule
        Epochs & Learning Rate \\ 
        \midrule
        \multicolumn{2}{c}{KITTI} \\
        \midrule
        0 to 2 & $5 \times 10^{-5}$ \\
        \midrule
        2 to 8 & $1 \times 10^{-4}$ \\
        \midrule
        8 to 20 & $1.5 \times 10^{-4}$ \\
        \midrule
        20 to 30 & $1 \times 10^{-4}$ \\
        \midrule
        30 to 45 & $5 \times 10^{-5}$ \\
        \midrule
        45 to 60 & $2 \times 10^{-5}$ \\
        \midrule
        \multicolumn{2}{c}{VOID} \\
        \midrule
        0 to 10 & $1 \times 10^{-4}$ \\
        \midrule
        10 to 15 & $5 \times 10^{-5}$ \\
        \midrule
        \multicolumn{2}{c}{NUYv2} \\
        \midrule
        0 to 10 & $1 \times 10^{-4}$ \\
        \midrule
        10 to 15 & $5 \times 10^{-5}$ \\
        \midrule
    \end{tabular}
    \caption{
        \textit{Learning schedule for KITTI, VOID, and NYUv2.}
    }
\label{tab:learning_schedule}
\vspace{-0.0em}
\end{table}

\begin{table}[t]
    \centering
    \footnotesize
    \setlength\tabcolsep{20pt}
    \begin{tabular}{l l l}
        \midrule
        Dataset & Min Pool & Max Pool \\ 
        \midrule
        KITTI \cite{uhrig2017sparsity}
        & 5, 7, 9, 11, 13 & 15, 17 \\
        \midrule
        VOID \cite{wong2020unsupervised}
        & 15, 17 & 23, 27, 29 \\
        \midrule
        NYUv2 \cite{silberman2012indoor}
        & 15, 17 & 23, 27 \\
        \midrule
    \end{tabular}
    \caption{
        \textit{Min pool and max pool kernel sizes for our sparse-to-dense module} Kernel sizes for VOID \cite{wong2020unsupervised} and NYUv2 \cite{silberman2012indoor} are larger because the point cloud generated from VIO \cite{fei2019geo} is much sparser than that of LIDAR in KITTI \cite{uhrig2017sparsity}.
    }
\label{tab:sparse_to_dense_kernel_sizes}
\vspace{-0.5em}
\end{table}

\begin{figure*}[t]
    \centering
    \includegraphics[width=\textwidth]{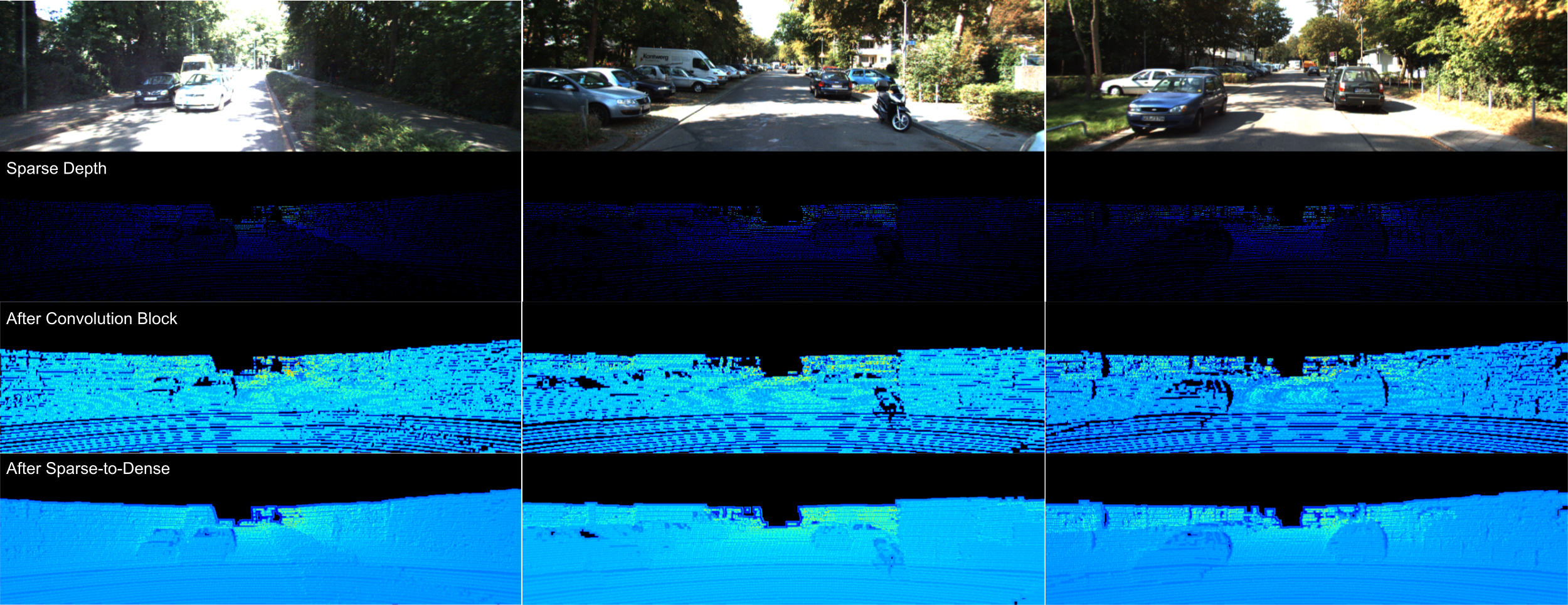}
    \caption{\textit{Visualization of depth features.} Row 3: ``After Convolution Block'' denotes the depth features produced by a typical first convolutional block used by \cite{ma2019self,wong2020unsupervised,yang2019dense} \textit{without} any form of densification. Row 4: ``After Sparse-to-Dense'' denotes the depth features learned by the proposed sparse-to-dense (S2D) module. Those learned without our module are still sparse; whereas S2D produces a dense or quasi-dense representation before it reaches the depth completion network. This alleviates the network from having to densify or propagate the sparse signal, making the overall architecture more efficient.}
    \label{fig:depth_features}
    \vspace{-0em}
\end{figure*}

\subsection{Implementation and Training Details}
We optimized our networks using Adam \cite{kingma2015adam} with $\beta_1=0.9$ and $\beta_2=0.999$. We trained for a total of 60 epochs on KITTI \cite{uhrig2017sparsity}, 15 epochs on VOID \cite{wong2020unsupervised}, and 15 epochs on NYUv2 \cite{silberman2012indoor}. We use a batch size of 8 with $768 \times 320$ crops for KITTI, $640 \times 480$ for VOID and $576 \times 416$ for NYUv2. For KITTI, we choose $w_{ph} = 1$, $w_{co}=0.15$, $w_{st}=0.95$, $w_{sz} = 0.6$, and $w_{sm} = 0.04$; for VOID and NYUv2, we set $w_{sz} = 2$ and $w_{sm} = 2$. Kernel sizes for our sparse-to-dense (S2D) module are shown in \tabref{tab:sparse_to_dense_kernel_sizes} for each dataset. We detail our learning rate schedule for each dataset in \tabref{tab:learning_schedule}. For data augmentations on KITTI, we performed random horizontal shifts to the image and depth map and randomly removed between 60\% to 70\% of the sparse points. For VOID and NYUv2, we randomly removed 30\% to 60\% of the sparse points. Augmentations are enabled 100\% of the time up for VOID and NYUv2. For KITTI it is applied 100\% of the time up to the 50th epoch and decreased by half every 5 epoch up to 60 epochs. Each augmentation has a 50\% probability of being applied.

\section{Features Learned by Sparse-to-Dense}
\label{sec:features_learned_sparse_to_dense}
In Sec 2.1 of the main text, we proposed a sparse-to-dense module (S2D) to learn a dense or quasi-dense representation of the sparse depth inputs. S2D utilizes a series of min and max pooling layers of various kernel sizes to densify the sparse depth inputs (for a list of kernel sizes used for each dataset, please see \tabref{tab:sparse_to_dense_kernel_sizes}). To balance the trade-off between density and detail (large vs. small kernel sizes), and near and far structures (min vs. max pooling), we concatenate the pooled results and learn three  $1 \times 1$ convolutions. The output of which is fused with the input sparse depth using a $3 \times 3$ convolution to ``fill in the gaps''.

\figref{fig:depth_features} shows visualizations of features learned by S2D and a comparison to the features learned by typical convolutional e.g. ResNet or VGG blocks used by \cite{ma2019self,wong2020unsupervised,yang2019dense}. Row 2 of \figref{fig:depth_features} shows that despite passing through several convolutional layers ($\approx 10K$ to $20K$ parameters), the representation obtained by a typical convolution block is still sparse; so the later layers will still have many zero-activations and must continue to densify the features. In contrast, using our proposed S2D ($\approx 900$ parameters), the depth representation learned is dense or quasi-dense before reaching the depth completion network (row 3). This enables non-zero activations in the later layers, which allows the network to use its early convolutions for learning scene geometry rather than densification. 

We note that our sparse-to-dense module may bare some resemblance to Spatial Pyramid Pooling (SPP) employed in classification \cite{he2015spatial} or stereo matching \cite{chang2018pyramid}. However, we note that \cite{he2015spatial} used SPP with max pooling to ensure that feature map sizes are consistent for different input sizes. \cite{chang2018pyramid} used average pooling to increase receptive field. Both use cases are intended for dense input. We discussed the drawbacks of max pooling \cite{he2015spatial} in Sec. 2.1 of the main text and showed in Table 3 of main text that SPP underperforms compare to our S2DM. Also we note that using average pooling  \cite{chang2018pyramid} will destroy the signal because the kernel will convolve and average over mostly zeros. The work that is most similar to our S2D module is the SPP for depth completion proposed by \cite{wong2021learning}. However, \cite{wong2021learning} only uses max pooling which decimates the detail of nearby structures. 

\section{Sensitivity Studies}
\label{sec:sensitivity_studies}
In this section, we provide additional studies to quantify the sensitivity of our model to incorrect calibration and various sparse depth density levels.

\begin{figure*}[t]
    \centering
    \includegraphics[width=1\textwidth]{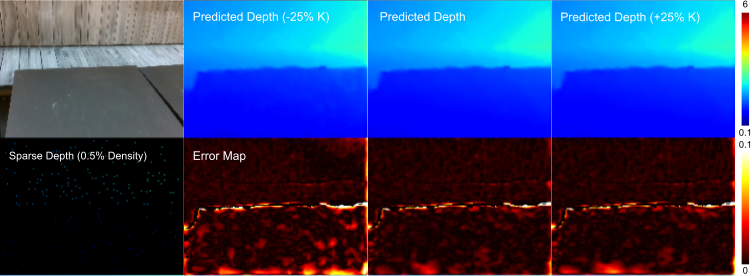}
    \caption{\textit{Visualization of predicted depth for incorrect calibration.} -25\% $K$ denotes 25\% decrease to intrinsics parameters and +25\% $K$ denotes 25\% increase.  Overall error in -25\% is increased (slight brigher shade of red). Larger errors caused by incorrect intrinsics is generally located at the edge of the depth map. +25\% have little effect on our predictions. This is because decreasing focal length causes surfaces to be distorted, which in turn affect depth predictions. On the other hand, increasing focal length packs points closer together, which is less detrimental in comparison.}
    \label{fig:void_test_set_calibration}
\vspace{-0em}
\end{figure*}

\begin{figure}[t]
    \centering
    \includegraphics[width=0.40\textwidth]{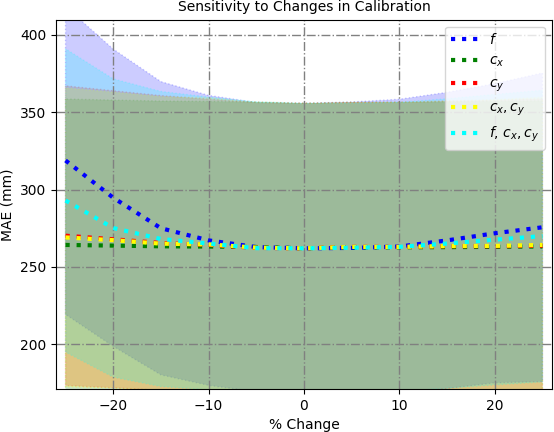}
    \caption{
        \textit{Sensitivity to changes in calibration on KITTI.} Focal length and principal point are altered to test sensitivity to changes in intrinsics parameters. Our method is robust to change up to $\approx10\%$ change. After which, performance degrades. We note that changes in principal point $(c_x, c_y)$ have little effect; whereas decreasing focal length ($f$) causes large drop in performance.
    }
    \label{fig:kitti_validation_set_calibration_sensitivity}
    \vspace{-0em}
\end{figure}

\subsection{To Incorrect Calibration}
\label{sec:sensitivity_incorrect_calibration}
\begin{figure}[t]
    \centering
    \includegraphics[width=0.40\textwidth]{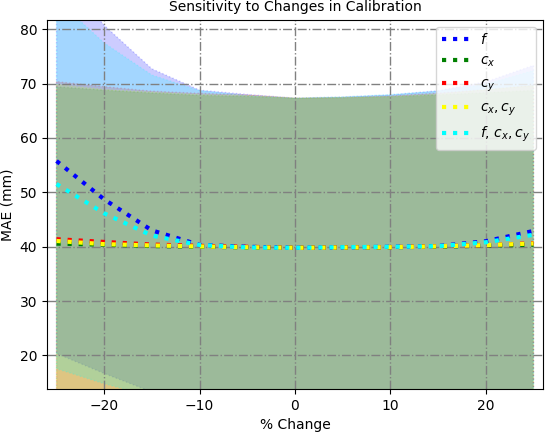}
    \caption{
        \textit{Sensitivity to changes in calibration on VOID.} Focal length and principal point are altered to test sensitivity to changes in intrinsics parameters. Our method is robust to change up to $\approx10\%$ change. After which, performance degrades. We note that changes in principal point $(c_x, c_y)$ have little effect; whereas decreasing focal length ($f$) causes large drop in performance.
    }
    \label{fig:void_test_set_calibration_sensitivity}
    \vspace{-0em}
\end{figure}

We showed in Table 5 of the main text that our method generalizes well when given the \textit{correct} calibration at test time. To consider the scenario of a miscalibrated camera, we studied the sensitivity of our model to \textit{incorrect} calibration on the KITTI dataset (outdoor scenarios) in Fig. 5 in the main text (also here in \figref{fig:kitti_validation_set_calibration_sensitivity}). Now, we further extend the sensitivity study to the indoor setting by conducting a similar sensitivity study on the VOID dataset (\figref{fig:void_test_set_calibration_sensitivity}). To this end, we consider changes to the focal length ($f$) and principal point $(c_x, c_y)$ parameters to create erroneous intrinsic calibration matrices for input to a pretrained model on VOID. 

The overall trend for indoor setting, (VOID, \figref{fig:void_test_set_calibration_sensitivity}) is similar to that of outdoor setting (KITTI, \figref{fig:kitti_validation_set_calibration_sensitivity}). For both indoors and outdoors, our model is robust to changes in principal point parameters $(c_x, c_y)$ -- increasing or decreasing them by up to 25\% has little effect on performance. This is because these parameters shifts the optical center so they do not affect the overall structure of the scene. We note that for large values outside of reasonable perturbation range will cause the performance to decrease. 

\begin{figure*}[h]
    \centering
    \includegraphics[width=1\textwidth]{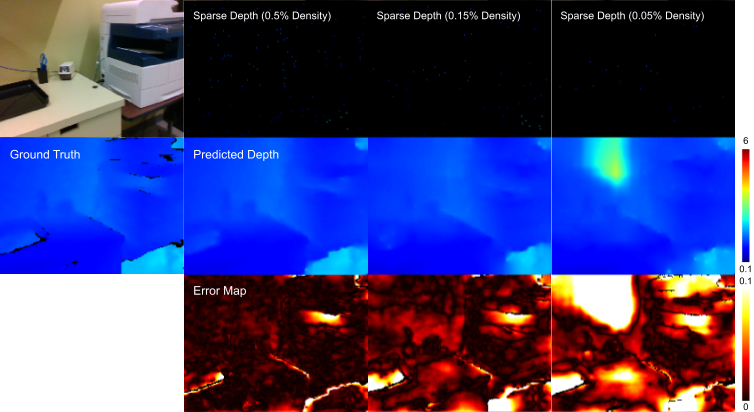}
    \caption{\textit{Visualization of predicted depth for various density levels on VOID.} Columns 1, 2: Our method works well for density levels of 0.5\% and 0.15\%. Column 3: The quality of predicted depth begins to degrade in far homogeneous regions where there are no sparse points e.g. wall when density level drops to 0.05\%. }
    \label{fig:void_test_set_density}
\vspace{-1em}
\end{figure*}

\begin{figure}[h]
    \centering
    \includegraphics[width=1\linewidth]{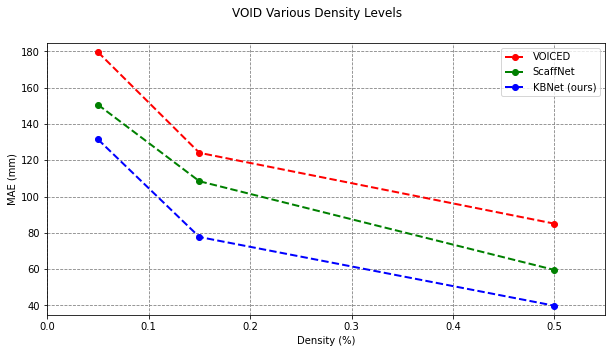}
    \caption{\textit{VOID test set across various density levels.} We compare our method against VOICED \cite{wong2020unsupervised} and ScaffNet \cite{wong2021learning} on the VOID test set for various density levels (0.5\%, 0.15\%, 0.05\%). In terms of MAE, our method performs better than other methods across all density levels.}
    \label{fig:void_test_set_density_mae_plot}
\vspace{-1em}
\end{figure}

\begin{table}
    \footnotesize
    \centering
    \setlength\tabcolsep{10pt}
    \begin{tabular}{l c c c c}
        \midrule 
        Method & MAE & RMSE & iMAE & iRMSE \\ \midrule
        \multicolumn{5}{c}{0.50\% Density} \\
        \midrule
        VOICED \cite{wong2020unsupervised} 
        & 85.05 & 169.79 & 48.92 & 104.02 \\ 
        \midrule
        ScaffNet \cite{wong2021learning} 
        & 59.53 & 119.14 & 35.72 & 68.36 \\ 
        \midrule
        Ours
        & \textbf{39.80} & \textbf{95.86} & \textbf{21.16} & \textbf{49.72} \\
        \midrule
        \multicolumn{5}{c}{0.15\% Density} \\
        \midrule
        VOICED \cite{wong2020unsupervised} 
        & 124.11 & 217.43 & 66.95 & 121.23 \\
        \midrule
        ScaffNet \cite{wong2021learning} 
        & 108.44 & 195.82 & 57.52 & 103.33 \\
        \midrule
        Ours
        & \textbf{77.70} & \textbf{172.49} & \textbf{38.87} & \textbf{85.59} \\
        \midrule
        \multicolumn{5}{c}{0.05\% Density} \\
        \midrule
        VOICED \cite{wong2020unsupervised} 
        & 179.66 & 281.09 & 95.27 & 151.66 \\
        \midrule
        ScaffNet \cite{wong2021learning} 
        & 150.65 & \textbf{255.08} & 80.79 & 133.33 \\
        \midrule
        Ours
        & \textbf{131.54} & 263.54 & \textbf{66.84} & \textbf{128.29} \\
        \midrule
    \end{tabular}
    \caption{
        \textit{Sensitivity study on various sparse depth density levels on VOID.} We train a single model on VOID using sparse depth maps of 0.50\% density and evaluate it on 0.50\%, 0.15\%, 0.05\% density test sets. As expected, performance degrade as the input become more sparse. Overall, we perform better than \cite{wong2021learning,wong2020unsupervised}; however, at 0.05\%, \cite{wong2021learning} performs better on the RMSE metric. 
    }
    \vspace{-2em}
\label{tab:void_test_density_study}
\end{table}

\begin{figure*}[th]
    \centering
    \includegraphics[width=\textwidth]{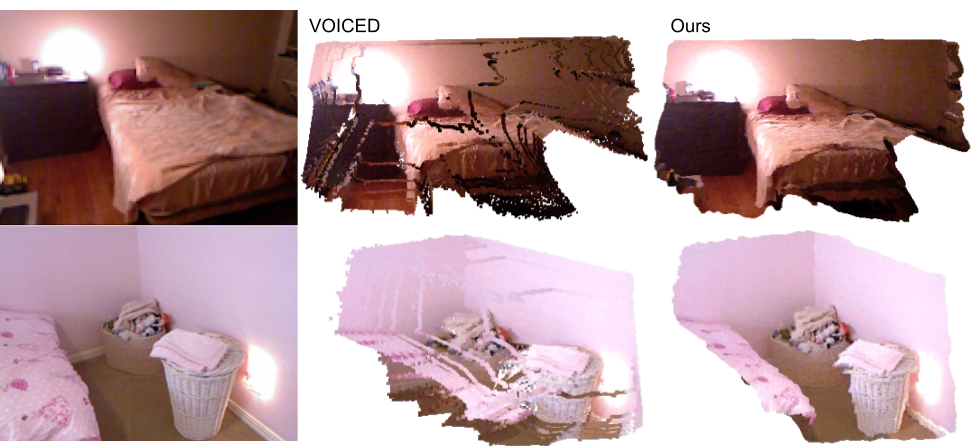}
    \caption{\textit{Qualitative results on generalization to novel scenes captured by a different sensor platform}. We trained our model on VOID \cite{wong2020unsupervised} (captured by Intel RealSense) and tested the model on NYUv2 \cite{silberman2012indoor} (captured by Microsoft Kinect). We also used a pretrained model (on VOID) of \cite{wong2020unsupervised} as the baseline and tested it on NYUv2. Here, we show the predicted depth as point clouds, backprojected to 3D and colored. \cite{wong2020unsupervised} predicted a distorted scene where the points are bowed towards the camera; whereas, while our predictions are not perfect, they are reasonable.}
    \label{fig:void_to_nyuv2_head_to_head}
\vspace{-0.5em}
\end{figure*}

Unlike its behavior with changes in the principal point, the model degrades when focal length ($f$) is decreased. For both indoors and outdoors, we are robust up to 10\% decrease in focal length, after which error will increase. We note that the performance drop is asymmetric, our model is robust to increases in focal length up to 20\%. The reason for this phenomeon is as follows: Geometrically, decreases in focal length will cause points to backproject to a wider field of view, which distorts surfaces by pushing points that belong to the same surface far from each other. On the other hand, increases in focal length will cause points to pack tighter together. This does not disrupt the scene structure for small values, but for large values, points will get squashed together; this is demonstrated by the small uptick in error when increasing focal length by 20 to 25\%.

We note that these values are well out of the typical range of calibration error and should not be of concern. For example when using off-the-shelf calibration packages that implements \cite{zhang2000flexible} to calibrate our camera, we obtained a standard error of $\approx 0.6\%$, which yields $\pm \approx 1.1\%$ margin of error for a 95\% confidence interval. Nonetheless, there exists the risk of using the wrong calibration; however, we believe this trade-off is well worth the performance boost provided by the proposed architecture.

\figref{fig:void_test_set_calibration} shows a visualization of depth predicted by our model when using erroneous calibration. -25\% $K$ denotes a 25\% decrease to focal length and principal point and +25\% $K$ denotes a 25\% increase to both. As we can see, the larger errors are typically located along the border of the predicted depth map; there is also a slight increase in error (brighter shade of red) for the entire scene. Increasing intrinsics by 25\% affects the output less significantly, but nonetheless we observe an increase in errors.

\subsection{To Various Density Levels}
\label{sec:sensitivity_density_levels}
In \tabref{tab:void_test_density_study}, we consider three different levels of density for the sparse depth inputs, 0.50\%, 0.15\%, 0.05\% of the image space, that are provided by the VOID dataset \cite{wong2020unsupervised}. To this end, we train a \textit{single} model on VOID using sparse depth maps of 0.50\% density and evaluate it on 0.50\%, 0.15\%, 0.05\% density test sets. We also compare our method against \cite{wong2021learning,wong2020unsupervised} under these density levels.

\begin{figure*}[th]
    \centering
    \includegraphics[width=\textwidth]{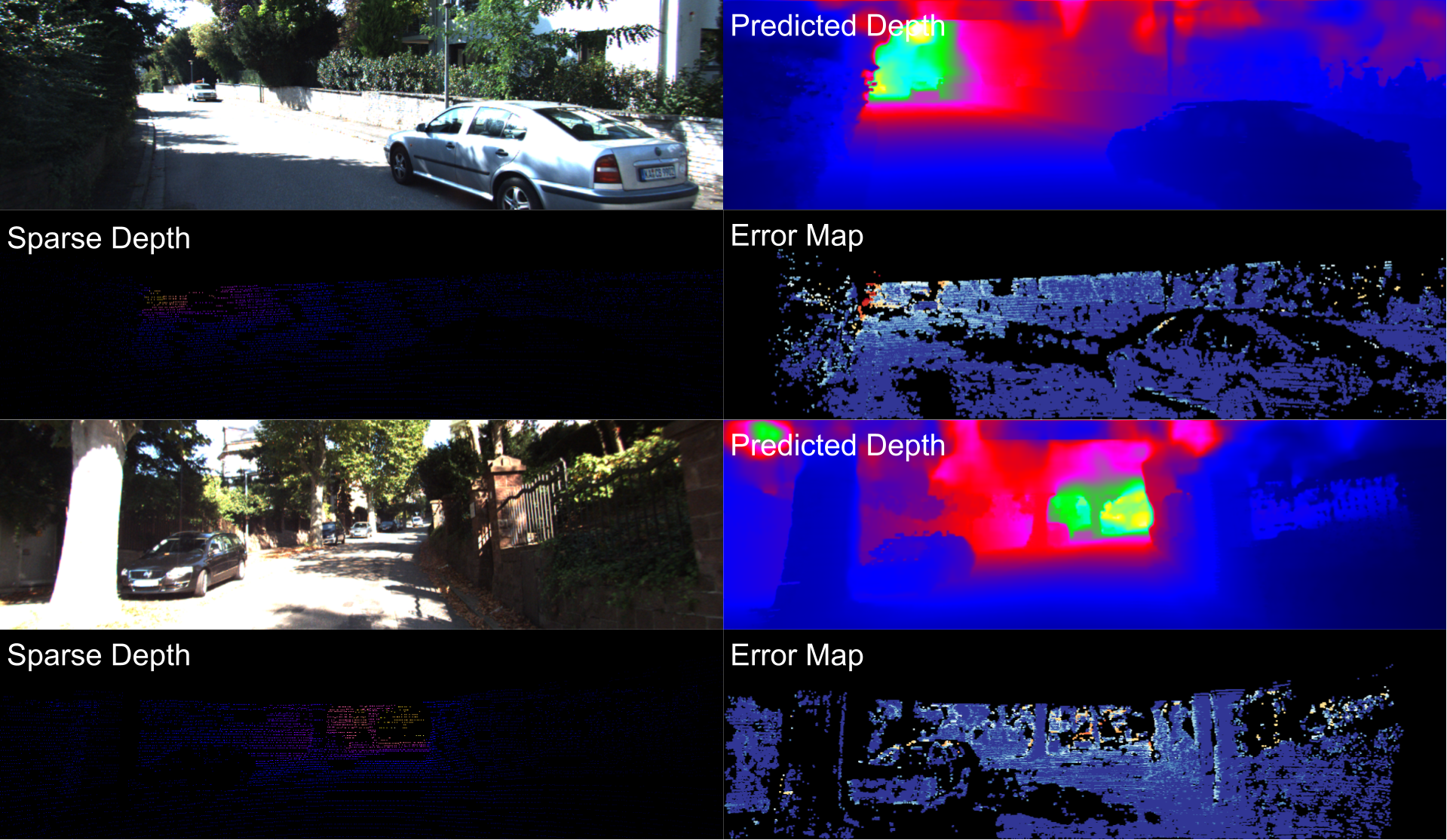}
    \caption{\textit{Qualitative results on KITTI depth completion benchmark.}}
    \label{fig:kitti_test_set_extra}
\vspace{-1em}
\end{figure*}

As expected, as density decreases, our performance also degrades. However, we still outperform both \cite{wong2021learning,wong2020unsupervised} under all three levels, see \figref{fig:void_test_set_density_mae_plot}. We note that at the sparsest setting of 0.05\%, \cite{wong2021learning} does beat us on the RMSE metric. The reason for this is that we selected the kernel sizes for our model based on the sparsity level of 0.5\%; therefore, when testing it on $10 \times$ sparser point cloud, our depth representation will be more sparse as well, which limits the potential of our calibrated backprojection layers. In contrast, \cite{wong2021learning} proposed a network to first estimate the dense coarse topology. This phenomenon is also observed in KITTI, shown in Table 3 of the main text, where we removed our sparse-to-dense module and we observed a significant drop in performance. 

\figref{fig:void_test_set_density} shows qualitative evaluations on the three density levels. For 0.50\%, error is low overall and the shape of the recovered scene resembles that of the ground truth. When we decrease density to 0.15\%, we observe slight blurring in object shapes and increased errors in homogeneous regions. At 0.05\%, we begin to observe artifact such as the green ``blob'' corresponding to the wall with more exaggerated errors in homogeneous regions. This is because locally the textureless surfaces give little to no information on object shape. Without sparse depth to anchor their values, they can be arbitrary. In this case the ``empty'' region is predicted as far.

\begin{table}[t]
    \footnotesize
    \centering
    \setlength\tabcolsep{5.5pt}
    \begin{tabular}{l c c c c}
        \midrule 
        Method & MAE & RMSE & iMAE & iRMSE \\ 
        \midrule
        ADNN \cite{chodosh2018deep}
        & 439.48 & 1325.37 & 3.19 & 59.39 \\ 
        \midrule
        Morph-Net \cite{dimitrievski2018learning}
        & 310.49 & 1045.45 & 1.57 & 3.84 \\
        \midrule
        CSPN \cite{cheng2018depth}
        & 279.46 & 1019.64 & 1.15 & 2.93 \\
        \midrule 
        KBNet (Ours)
        & \textit{256.76} & \textit{1069.47} & \textit{1.02} & \textit{2.95} \\
        \midrule
        SS-S2D \cite{ma2019self} 
        & 249.95 & 814.73 & 1.21 & 2.80	\\ 
        \midrule
        DeepLiDAR \cite{qiu2019deeplidar}
        & 226.50 & 758.38 & 1.15 & 2.56	 \\ 
        \midrule
        PwP \cite{xu2019depth}
        & 235.73 & 785.57 & 1.07 & 2.52 \\ 
        \midrule			
        UberATG-FuseNet \cite{chen2019learning}   
        & 221.19 & 752.88 & 1.14 & 2.34 \\ 
        \midrule
        RGB\_guide\&certainty \cite{van2019sparse}
        & 215.02 & 772.87 & 0.93 & 2.19	\\ 
        \midrule
        DDP \cite{yang2019dense}
        & 203.96 & 832.94 & 0.85 & 2.10 \\ 
        \midrule		
        CSPN++ \cite{cheng2020cspn++} 
        & 209.28 & 743.69 & 0.90 & 2.07 \\ 
        \midrule
        NLSPN \cite{park2020non} 
        & \textbf{199.59} & \textbf{741.68} & \textbf{0.84} & \textbf{1.99}  \\ 
        \midrule
    \end{tabular}
    \caption{
        \textit{KITTI supervised depth completion benchmark.} Results are directly taken from online leaderboard. Note: SS-S2D \cite{ma2019self} and DDP \cite{yang2019dense} compete in both supervised and unsupervised benchmarks. Our results are italicized. Despite being an unsupervised method, our method beats some supervised methods \cite{chodosh2018deep,dimitrievski2018learning} and our iMAE score (1.02) is ranked 5th amongst supervised methods. 
    }
    \vspace{-2em}
\label{tab:kitti_supervised_benchmark}
\end{table}

\begin{table}[t]
    \footnotesize
    \centering
    \setlength\tabcolsep{9.5pt}
    \begin{tabular}{l c c c c}
        \midrule
        Method & MAE & RMSE & iMAE & iRMSE \\ 
        \midrule
        SGDU \cite{schneider2016semantically}
        & 605.47 & 2312.57 & 2.05 & 7.38 \\ 
        \midrule
        SS-S2D \cite{ma2019self}
        & 350.32 & 1299.85 & 1.57 & 4.07 \\ 
        \midrule
        IP-Basic \cite{ku2018defense} 
        & 302.60 & 1288.46 & 1.29 & 3.78 \\ 
        \midrule
        DFuseNet \cite{shivakumar2019dfusenet}
        & 429.93 & 1206.66 & 1.79 & 3.62 \\ 
        \midrule
        DDP* \cite{yang2019dense}
        & 343.46 & 1263.19 & 1.32 & 3.58 \\ 
        \midrule
        VOICED \cite{wong2020unsupervised}  
        & 299.41 & 1169.97 & 1.20 & 3.56 \\ 
        \midrule
        AdaFrame \cite{wong2021adaptive}
        & 291.62 & 1125.67 & 1.16 & 3.32 \\ 
        \midrule
        SynthProj* \cite{lopez2020project}
        & 280.42 & 1095.26 & 1.19 & 3.53 \\
        \midrule
        ScaffNet* \cite{wong2021learning}  
        & 280.76 & 1121.93 & 1.15 & 3.30 \\ 
        \midrule
        KBNet (Ours)	
        & \textit{256.76} & \textit{1069.47} & \textit{1.02} & \textit{2.95} \\
        \midrule
    \end{tabular}
    \caption{
        \textit{KITTI unsupervised depth completion benchmark}. Results are directly taken from online leaderboard. Note: SS-S2D \cite{ma2019self} and DDP \cite{yang2019dense} compete in both supervised and unsupervised benchmarks. Our method outperforms is trained only on KITTI, but still the state of the art \cite{wong2021learning} (trained on KITTI and Virtual KITTI \cite{gaidon2016virtual}) by an average of 8.8\% across all metrics. * denotes methods that use additional synthetic data for training.
    }
    \vspace{-1em}
\label{tab:kitti_unsupervised_benchmark}
\end{table}

\section{Generalization to Other Sensor Platforms}
\label{sec:generalization_to_other_sensor_platforms}
In Sec. 3.4 of the main text, we discussed our ability to generalize to other sensor platforms that may use a different test time camera than one used to collect training data. In Table 5 of the main text, we showed quantitatively that we generalize better than the baseline. Here, we demonstrate this qualitatively in \figref{fig:void_to_nyuv2_head_to_head}. 

To this end, we trained our model on VOID \cite{wong2020unsupervised} (captured by Intel RealSense) and tested the model on NYUv2 \cite{silberman2012indoor} (captured by Microsoft Kinect). We similarly trained the baseline \cite{wong2020unsupervised} on VOID and tested it on NYUv2. \figref{fig:void_to_nyuv2_head_to_head} shows the predicted depth, backprojected to the point clouds in 3D and colored. As we can see, \cite{wong2020unsupervised} predicted a distorted scene; in contrast, ours is not perfect, but reasonable. This demonstrates the benefit of taking calibration as input. It allows the model to generalize well when it is deployed to a sensor platform where the camera that is used is \textit{different} than the one used for training. We also note that neither models have been trained on NYUv2 which features a different scene distribution than that of VOID.

\section{KITTI Depth Completion Benchmark}
\label{sec:kitti_depth_completion_benchmark}
In Sec. 3.3 of the main text, we compare our method against \textit{unsupervised} methods on the KITTI online leaderboard. Here, we show quantitative comparisons against both supervised (\tabref{tab:kitti_supervised_benchmark}) and unsupervised (\tabref{tab:kitti_unsupervised_benchmark}) methods. Results and method names are directly taken from the KITTI online leaderboard. Here we refer to our method as KNBet, as listed on the leaderboard. We note that SS-S2D \cite{ma2019self} and DDP \cite{yang2019dense} compete in both supervised and unsupervised benchmarks. Additionally, we provide high resolution examples of our output in \figref{fig:kitti_test_set_extra}.

Despite being trained without ground-truth annotations, \tabref{tab:kitti_supervised_benchmark} shows that our method is competitive even amongst supervised method. We outperform some supervised methods \cite{cheng2018depth, chodosh2018deep, dimitrievski2018learning} across most metrics. We note that our method significantly improves the iMAE and iRMSE metrics, to the point where we are comparable to some of the supervised methods for close range performance. Our iMAE score, which penalizes mean error in close range regions, is ranked 5th overall amongst both supervised and unsupervised methods. To the best of our knowledge, we are the first work in unsupervised depth completion to demonstrate comparable performance to supervised methods. We note that supervised methods are generally more computationally expensive with high model complexity e.g. in terms of number of parameters, \cite{park2020non} uses 25.84M, \cite{qiu2019deeplidar} 53.4M, and \cite{xu2019depth} 28.99M; whereas we only use 6.9M. 

Compared to unsupervised methods (\tabref{tab:kitti_unsupervised_benchmark}), we rank first amongst all methods with the best scores across all metrics. Our model even beat methods \cite{lopez2020project,yang2019dense,wong2021learning} that use additional synthetic data (Virtual KITTI \cite{gaidon2016virtual}) for training, amongst which is the state of the art \cite{wong2021learning}. Despite this, we beat \cite{wong2021learning} by an average of 8.8\% across all metrics while using 11.5\% fewer parameters. These results demonstrates the potential of our method to bridge the gap between supervised and unsupervised method. Moreover, our network is light-weight and can be deployed on VIO system \cite{fei2019geo}. While there is still a long road ahead, these results show a lot of promise in enabling unsupervised methods to learn online and to be used for real-time application for low-cost hardware systems. 

\end{document}